\documentclass[10pt,twocolumn,letterpaper]{article}

\usepackage[pagenumbers]{iccv} 

\usepackage[accsupp]{axessibility}
\usepackage{booktabs}
\usepackage{amsmath, amssymb}
\usepackage{tikz}
\usepackage[export]{adjustbox}
\usetikzlibrary{positioning}
\usepackage{multirow,bigdelim}

\definecolor{CBRed}{RGB}{204, 102, 119}
\definecolor{CBYellow}{RGB}{221, 204, 119}
\definecolor{CBBlue}{RGB}{136, 204, 238}
\definecolor{CBMagenta}{RGB}{170, 68, 153}
\definecolor{CBOrange}{RGB}{230, 159, 0}

\newcommand{\myparagraph}[1]{\vspace{3mm}\noindent\textbf{#1}~}

\newcommand{\change}[1]{\textcolor{black}{#1}}

\definecolor{iccvblue}{rgb}{0.21,0.49,0.74}
\usepackage[pagebackref,breaklinks,colorlinks,allcolors=iccvblue]{hyperref}
\usepackage{cuted}
\def\paperID{12828} 
\def\confName{ICCV}
\def\confYear{2025}

\title{DiffuMatch: Category-Agnostic Spectral Diffusion Priors for\\Robust Non-rigid Shape Matching}

\author{Emery Pierson$^{1}$ \hspace{1cm}
Lei Li$^{2}$ \hspace{1cm}
Angela Dai$^{2}$ \hspace{1cm}
Maks Ovsjanikov$^{1}$ \hspace{1cm} \\
LIX, Ecole Polytechnique$^{1}$ \hspace{1cm} Technical University of Munich$^{2}$
}

\begin{document}

\begin{figure}
\twocolumn[{%
\renewcommand\twocolumn[1][]{#1}%
\maketitle
\begin{center}
    \centering
    \vspace{-0.6cm}
    \includegraphics[width=0.8\linewidth]{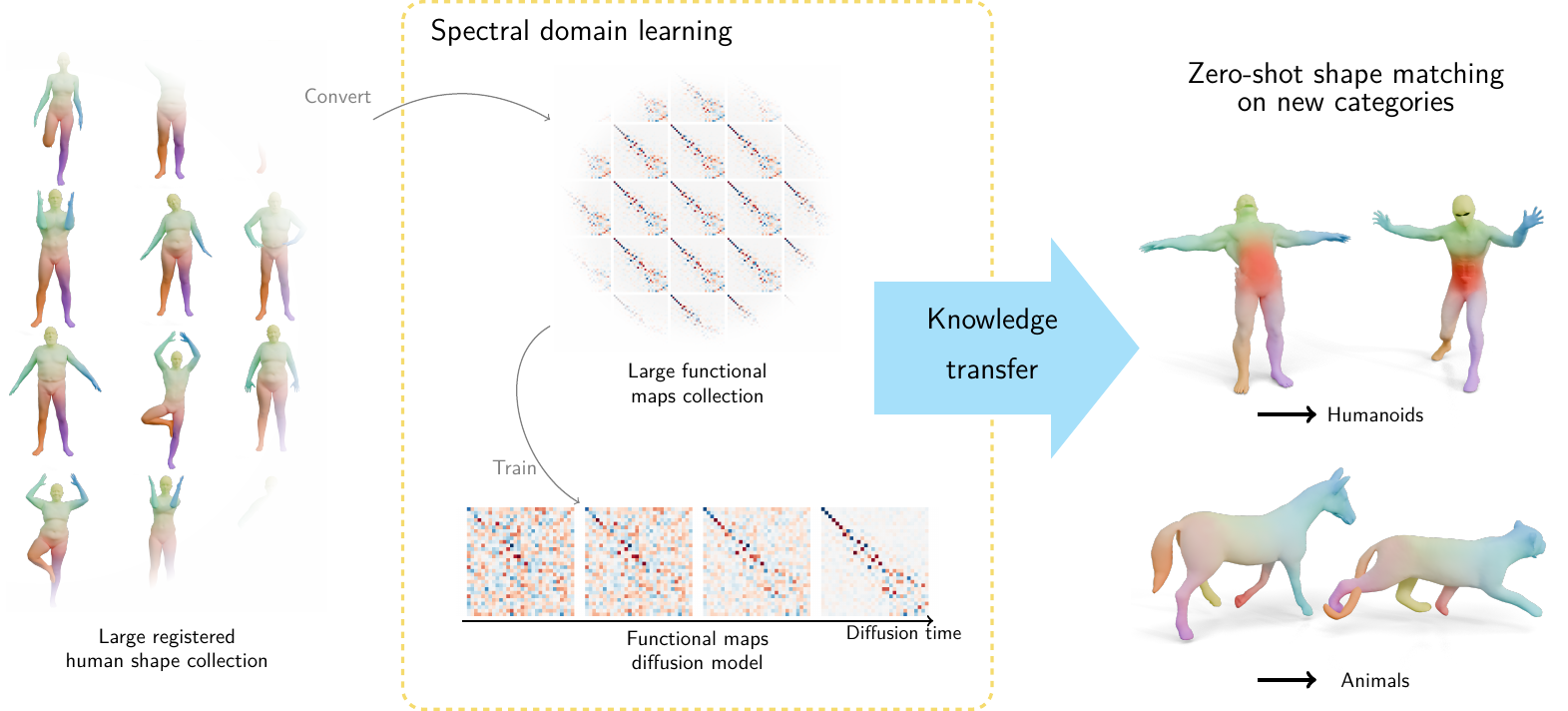}
    \vspace{-0.3cm}
    \caption{
    We learn diffusion priors in the spectral domain from a large collection of functional maps computed on registered human shapes. The learned spectral diffusion priors are category-agnostic and generalize robustly to unseen shape categories, enabling accurate zero-shot non-rigid shape matching.
    }
    \label{fig:teaser}
\end{center}%
}]
\end{figure}

\begin{abstract}
    Deep functional maps have recently emerged as a powerful tool for solving non-rigid shape correspondence tasks.  Methods that use this approach combine the power and flexibility of the functional map framework, with data-driven learning for improved accuracy and generality. However, most existing methods in this area restrict the learning aspect only to the feature functions and still rely on axiomatic modeling for formulating the training loss or for functional map regularization inside the networks. This limits both the accuracy and the applicability of the resulting approaches only to scenarios where assumptions of the axiomatic models hold. In this work, we show, for the first time, that both \emph{in-network regularization} and functional map training can be replaced with data-driven methods. For this, we first train a generative model of functional maps \emph{in the spectral domain} using score-based generative modeling, built from a large collection of high-quality maps. We then exploit the resulting model to promote the structural properties of ground truth functional maps on new shape collections. Remarkably, we demonstrate that the learned models are category-agnostic, and can fully replace commonly used strategies such as enforcing Laplacian commutativity or orthogonality of functional maps. Our key technical contribution is a novel distillation strategy from diffusion models in the spectral domain. Experiments demonstrate that our learned regularization leads to better results than axiomatic approaches for zero-shot non-rigid shape matching. Our code is available at: \url{https://github.com/daidedou/diffumatch/}  

\end{abstract}

\section{Introduction}
\label{sec:Introduction}

Shape matching is a fundamental problem in geometry processing, as it is a necessary step for many applications such as 
shape interpolation \cite{alexa2023rigid}, texture transfer \cite{dinh2005texture}, and statistical shape analysis \cite{bogo2014faust,bogo2017dynamic}.

A particularly appealing approach to non-rigid shape matching is the recent deep functional maps framework~\cite{litany2017deep, donati2020deep}. It consists of two main blocks: (1) a deep feature extractor that computes descriptor functions approximately preserved across a pair of input shapes,
and (2) a differentiable functional map solver that computes the matching in the spectral domain under axiomatic regularizations.
Functional maps~\cite{funcmaps} provide a flexible and compact representation of shape correspondences as small-sized matrices, and have been successfully applied to shape matching, but also other tasks such as shape classification~\cite{halimi2018self} or representation alignment~\cite{fumerolatent}.
However, existing methods rely heavily on axiomatic modeling, such as near isometry and local area preservation constraints \cite{Cao2023UnsupervisedLO,sharma2020weakly}, for formulating the training loss or for functional map regularization inside the networks.
This limits their accuracy and generalization to unseen shape pairs where these assumptions may not hold.

This motivates us to explore replacing the axiomatic regularizations in deep functional maps with structural priors learned directly from data.
Recently, diffusion models \cite{song2019generative,ho2020denoising} have demonstrated strong capabilities in modeling complex data distributions, achieving impressive results in tasks such as image generation and editing~\cite{rombach2022high,meng2021sdedit}. 
The rich priors learned by these models have also proven useful in many other tasks, such as 3D generation and reconstruction \cite{poole2023dreamfusion,wang2024prolificdreamer,Wu_2024_CVPR}.
Inspired by these successes, we propose to leverage diffusion models to capture structural properties of functional maps directly in the spectral domain. With the spectral diffusion priors as a powerful regularizer, we aim to enhance the robustness and accuracy of functional map estimation for non-rigid shape matching.


In this work, we leverage large-scale datasets of registered non-rigid 3D shapes, primarily human bodies \cite{bogo2017dynamic},
to learn informative structural priors of functional maps. 
We first construct a large collection of high-quality functional maps from the non-rigid 3D registrations in \cite{bogo2017dynamic}, and then train an unconditional diffusion model in the spectral domain to capture the distribution of these functional maps.
Building on this spectral diffusion model, we propose a novel zero-shot deep functional map pipeline, which incorporates a novel data-driven regularization to promote structural properties consistent with those observed in the training data.
Specifically, we distill a mask from the trained spectral diffusion model.
This mask encodes learned structural priors, replacing conventional axiomatic regularizations, such as Laplacian commutativity or orthogonality, commonly used in deep functional map pipelines.
Remarkably, our learned spectral diffusion priors, though trained on human shapes, are category-agnostic and demonstrate strong generalization to unseen shape categories, including humanoids and animals.


In summary, our contributions are: 
\begin{itemize}
    \item We introduce a spectral diffusion model to learn the distribution of functional maps, effectively capturing their structural characteristics in the spectral domain.
    \item We distill the learned spectral diffusion priors into a mask that serves as a data-driven regularizer, replacing axiomatic regularizations and improving robustness in zero-shot deep functional map pipelines.
    \item Our spectral diffusion priors, learned from human shapes, demonstrate remarkable adaptability, generalizing effectively to diverse and previously unseen shape categories.
\end{itemize}


\section{Related Work}
\label{sec:RelatedWork}

\myparagraph{Functional Maps and Regularizations.}
Since the functional maps seminal work~\cite{funcmaps}, in which orthogonality and Laplacian commutativity penalties are derived to encourage near-isometric maps,  many axiomatic approaches have been proposed to improve the computation of functional maps. One of the most common penalties is to encourage bijectivity of the functional maps~\cite{ovsjanikov2016computing}, thereby encouraging bijectivity of the corresponding pointwise correspondence. 
Ren~\etal~\cite{ren2018continuous} propose to encourage orientation preservation and continuity of maps, along with a new iterative algorithm to improve the quality of maps. Panine~\etal~\cite{panine2022non} propose to promote conformality of maps with a new penalty and functional basis for the map computation.

Recently, Zoomout~\cite{zoomout} has shown that alternating between functional map and pointwise correspondence representation is a remarkably efficient method to regularize the final map quality automatically. One of the key components is the projection of maps to the proper map space~\cite{ren2021discrete,li2022learning}, the space of functional maps corresponding to valid point correspondences, from which it is easier to optimize spatial energies like the Dirichlet energy~\cite{magnet2022smooth}, \change{or elastic energies with a separate deformation network~\cite{cao2024spectral}. This has lately been used to improve the training of deep functional maps, with different losses encouraging functional maps to be proper~\cite{li2022srfeat,Cao2023UnsupervisedLO,cao2024synchronous}. Some approaches propose to use geometric information to refine maps using geometric consistency either in the training of deep functional maps~\cite{cao2024synchronous}, or at test time with precomputed deep features~\cite{gao2023sigma, roetzer2024discomatch, roetzer2024spidermatch}, which can be useful in partial shape matching~\cite {ehm2024partial, ehm2024geometrically}. Another way to mitigate errors is to match multiple shapes at the same time rather than a pair~\cite{eisenberger2023g, gao2021isometric}, with an increased computational cost. Another direction aims at exploring an alternative to the Laplace-Beltrami operator for constructing the shape basis~\cite{bracha2020corres, bensaid2023partial, gao2023sigma, weber2024finsler}, however, it is not straightforward to incorporate these new approaches in a deep functional maps pipeline.}

Only a few works have focused on improving mask regularization. In partial functional maps~\cite{rodola2017partial}, the authors propose a slanted regularization to encourage the map to follow Weyl's law. Ren~\etal{} improved the original Laplacian commutativity by using the Resolvent operator~\cite{ren2019structured}, which has better theoretical properties.

To perform well, most of these approaches require a good quality initialization as input, which is given by the mask regularization~\cite{attaiki2022ncp}, \change{or by large-scale pre-trained features~\cite{yona2025neural}}. In contrast, we propose a data-driven mask computation, which allows for a better initialization of the maps and a distilled loss to optimize the maps. 

\myparagraph{Knowledge Distillation of Score-based Generative Models.}
Denoising diffusion models~\cite{ho2020denoising} are a class of generative models that learn the mapping between an (unknown) data distribution and a Gaussian distribution. They have shown a great generalization capability, surpassing GANs for image generation~\cite{dhariwal2021diffusion}.  Moreover, the distribution learned by the diffusion models, has numerous desirable properties, as it learns the gradient of the log density, the score, of the (noised) data distribution~\cite{song2021scorebased}. The learned score can be distilled in various ways depending on the downstream tasks, such as image inpainting~\cite{lugmayr2022repaint}, denoising~\cite{song2022solving}, and more recently, text-to-3D generation.

Score distillation sampling (SDS)~\cite{poole2023dreamfusion} has indeed quickly become a preferred approach to zero-shot text-to-3D generation using 2D image-based diffusion models. The authors of~\cite{poole2023dreamfusion} propose to use the learned score of the diffusion model as the gradient of a desired image given a user text prompt. Coupled with a differentiable scene representation and rendering, the proposed loss allows for the accurate generation of new 3D scenes. However, the approach exhibits undesired properties, such as almost deterministic generation (convergence towards the mean image corresponding to the prompt), low-quality shapes, or color saturation of the generated shapes. To overcome this limitation, HiFA~\cite{zhu2024hifa} proposes a strategy to mitigate those effects by using negative prompts and forcing realistic images by emphasizing SDS steps with low levels of noise. ProlificDreamer~\cite{wang2024prolificdreamer} instead proposes to learn a fine-tuned diffusion model for the prompt, to avoid deterministic generation. Those approaches, however, require a conditional diffusion model to work properly. Lukoianov~\etal~\cite{lukoianov2024score} proposes DDIM inversion to follow the score to avoid wrong gradient directions. However, the inversion process is approximate and can be slow. 

Most methods presented here are designed for image generation. As we will see in \cref{sec:Method}, their generalization to the regularization of functional maps is not straightforward, and we therefore propose to adapt the distillation strategy for robust non-rigid shape matching.

\myparagraph{3D Generative Modeling for Shape Matching.}
Generative modeling has proven to be a powerful tool for solving shape-matching tasks. In 3D-CODED~\cite{groueix2018_3DCODED}, the authors train a point-cloud auto-encoder on a large synthetic human dataset and register human scans in a zero-shot approach. The auto-encoder approach has been improved with geometric regularization to avoid degenerated reconstructions. Neural Jacobian Fields~\cite{NJF_2022} improve the overall quality of reconstructions by predicting the Jacobian of deformations instead of directly predicting the deformation field, which implicitly regularizes the final shape. ARAPReg~\cite{ARAPREG_2021} learns a geometrically regularized latent space by penalizing directions that increase the ARAP energy. This strategy has improved to learn correspondences on unregistered datasets~\cite{yang2024gencorres} but requires a two-step training strategy. Finally, some works directly learn to generate the matching of shapes, whether directly in the spatial domain~\cite{eisenberger2020deep, trappolini2021shape, Merrouche_2023_BMVC} or in the spectral domain using deep functional maps~\cite{litany2017deep, halimi2019unsupervised}. In particular, deep functional methods have shown great success for intra-category training of shape matching approaches~\cite{roufosse2019unsupervised, li2022learning, cao2022unsupervised, Cao2023UnsupervisedLO, sun2023spatially}. A concurrent work~\cite{zhuravlev2025denoising} proposes to learn conditional distribution of functional maps along with shape descriptors. All the works mentioned above need training on specific categories before being used at test time: a model trained on humans is useful for registering other humans but often fails to generalize well to different categories, such as animals. 

We overcome this limitation by training an unconditional diffusion model on the space of functional maps. We propose a novel distillation strategy adapted to the specific problem of functional map regularization. Our approach generalizes to new categories of shapes (\eg{}, animals).

\section{Background \& Motivation}
\label{sec:Background}

\begin{figure*}[h]
    \centering
    \includegraphics[width=0.9\linewidth]{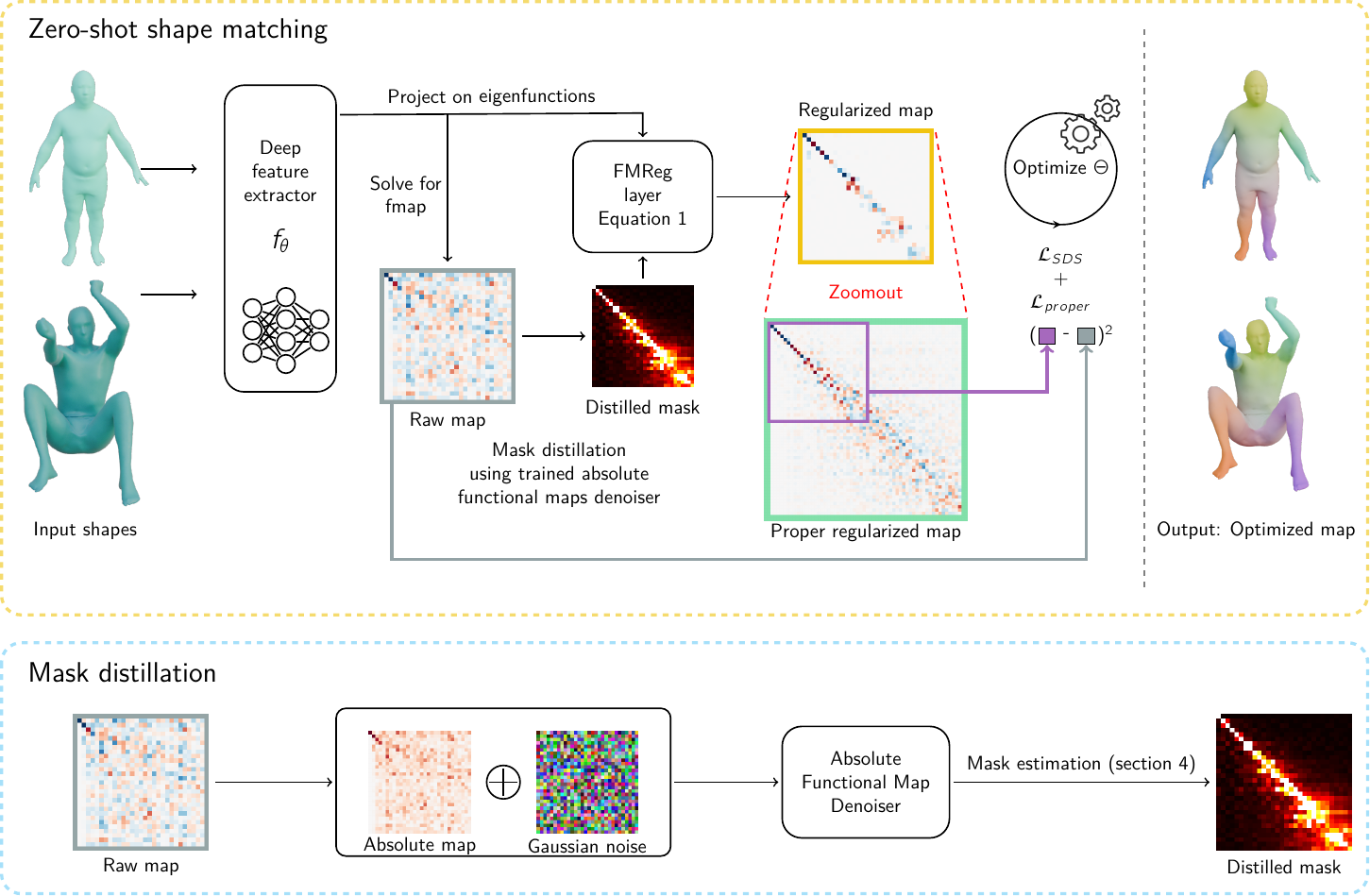}
    \vspace{-0.3cm}
    \caption{Our diffusion-based zero-shot shape matching pipeline. Given two shapes, we distill a mask using the spectral diffusion denoiser applied to the estimated ``raw'' functional map from features. We then use the distilled mask in a regularized functional map solver (FMReg) and apply Zoomout \cite{zoomout} to obtain a proper, regularized map. We minimize both score distillation and L2 distance to the proper map.}
    \label{fig:overview}
\end{figure*}

\subsection{Deep Functional Maps}\label{sec:funcmaps}

The objective of deep functional maps is to learn shape descriptors to compute high-quality correspondences on pairs of shapes $(\mathcal{S}_1, \mathcal{S}_2)$, represented as triangular meshes. Let $n_1, n_2$ be their respective number of vertices. The pipeline generally consists of the following steps:

\begin{itemize}
    \item Compute the first $k$ eigenfunctions of an intrinsic surface operator -- usually the Laplace-Beltrami operator -- on each shape, serving as a basis of functions on these shapes. The Laplacian is discretized as $S^{-1}W$, where $S$ is the diagonal matrix of mesh vertex areas,  and $W$ is the cotangent weight matrix. The eigenfunctions are stored as matrices in the form of $\Phi_1 \in \mathbb{R}^{n_1\times k}$ and $\Phi_2 \in \mathbb{R}^{n_2\times k}$.
    \item A set of $d$ descriptor functions (approximately preserved by the unknown map) 
    $F_1, F_2 = f_{\theta}(\mathcal{S}_1) \in \mathbb{R}^{n_1\times d}$, $ f_{\theta}(\mathcal{S}_2) \in \mathbb{R}^{n_2\times d}$ 
    extracted using a neural network $f_\theta: \mathcal{S} \mapsto \mathbb{R}^d$. After projecting them onto the respective eigenfunctions, the resulting descriptor coefficients are stored as matrices $A_1, A_2 \in \mathbb{R}^{k\times d}$, respectively.
    \item The functional map matrix $C$ between $\mathcal{S}_1$ and $\mathcal{S}_2$ is computed by solving the following:
    \begin{equation}
        C = \underset{C}{\text{ argmin }} ||CA_1 - A_2||^2 + \alpha ||M_{\text{reg}} C||^2,
        \label{eq:fmap_compute}
    \end{equation}
    where the first term is a data preservation term between descriptors, and the second term regularizes the map structure by using a sparsity promoting mask $M_{\text{reg}}$, derived from Laplacian or Resolvent operator commutativity. The whole pipeline $(\mathcal{S}_1, \mathcal{S}_2) \mapsto C$, also called FMReg layer~\cite{donati2020deep} is fully differentiable with respect to $\theta$. 
    \item The weights $\theta$ are optimized during training with axiomatic regularization terms, like area preservation, which is formulated as an orthogonality penalty:
    \begin{equation}
    \mathcal{L}_{\text{ortho}} (C) = ||CC^T - I||^2,
    \end{equation}
    or other regularizations, such as orientation preservation.
    \item A test-time, we use the learned pipeline to extract $C$. Post-processing algorithms such as Zoomout can be used to increase the accuracy of the map, before extracting the point-to-point-map using the aligned eigenfunctions $\Phi_1 C^T$ and $\Phi_2$ and nearest neighbor search or more accurate techniques~\cite{pai2021fast}.
\end{itemize}

\noindent Orthogonality and commutativity penalties are essential to ensure that the correspondences are plausible. We propose to replace those axiomatic penalties in deep functional maps with data-driven penalties by distilling priors from trained spectral diffusion models in a zero-shot manner. 

\subsection{Score-based Generative Modeling}

In this section, we follow the formalism of~\cite{karras2022elucidating} to present denoising score models. The general objective of generative modeling is to learn a distribution $p_\psi(x)$ (where $\psi$ are the learned parameters) corresponding to an (unknown) data distribution using the available samples of this distribution. Denoising score matching~\cite{hyvarinen2005estimation}, and in particular, denoising diffusion models, are a specific class of models that learn to model the score function, defined as:
\begin{equation}
    s(x) = \nabla_x \log p(x),
\end{equation}
instead of the density p. This formulation overcomes the problem of normalizing constants of the data density. Knowing the score function is equivalent to knowing the data distribution, as one can sample from it using Langevin dynamics~\cite{parisi1981correlation}. Since the score is unknown in parts of the sample space without data, denoising score matching learns the score functions $s_\theta(x_\sigma, \sigma) = \nabla_{x_\sigma} \log q(x_\sigma, \sigma)$ at different noise scales~\cite{vincent2011connection}, where $x_\sigma = x + n_\sigma$, with $n_\sigma \sim \mathcal{N}(o, \sigma^2I)$. This is done by learning a denoiser network $ D_\psi(x + n_\sigma, \sigma) $ by minimizing the following loss:

\begin{equation}
\mathbb{E}_{x \sim p_{\text{data}}} \mathbb{E}_{n_\sigma \sim \mathcal{N}(0, \sigma^2 I)}|| D_\psi(x + n_\sigma, \sigma) - x ||^2,
\end{equation}
\change{where the optimized parameters are the parameters $\psi$ of the denoiser.
We drop the denoiser parameters sign $\psi$ to avoid confusion with other learnable parameters, since for the rest of the paper, the denoiser is considered as trained.} 
After training, new samples are generated by progressively denoising random samples, following a probability ordinary differential equation~\cite{song2021scorebased}. Moreover, the score at noise level $\sigma$ can be estimated using:
\begin{equation}
\nabla_{x_\sigma} \log p({x_\sigma}; \sigma) = (D({x_\sigma}; \sigma) - x)/\sigma^2
\label{eq:score}
\end{equation}

\myparagraph{Score Distillation Sampling.}
\change{Score distillation sampling (SDS) \cite{poole2023dreamfusion} is a generic way of transferring knowledge from a diffusion model learned on a \textit{source domain} $\Omega$, to regularize or generate samples $y$ in a \textit{target domain}. It can be summarized as follows:
(1) obtain a trained denoiser $D(x_\sigma, \sigma)$, with $x \in \Omega$, the \textit{source domain}, on which it is easy to train the denoiser;
(2) differentiably extract $x = g(y_\theta)$ from the \textit{target} to the \textit{source} domain, where the representation $y_\theta$ of samples in the target domain is optimizable, and $g$ is a differentiable mapping from the \textit{target} to the \textit{source} domain.}
\change{In the original work \cite{poole2023dreamfusion}, the \textit{source domain} is images, and the \textit{target domain} is 3D scenes. 3D scenes are parameterized using Neural Radiance Fields $y_\theta$, and the function $g(y_\theta)$ is simply the differentiable rendering of a novel view.}


At each iteration, SDS consists in sampling $x=g(y_\theta) \in \Omega$ and perturbing the $x$ with noise $n_\sigma \sim \mathcal{N}(0, \sigma)$. Then, SDS guides the target representation $\psi$ by applying the following gradient to the parameters:
\begin{equation}
\nabla_\theta \mathcal{L}_{\text{SDS}} = \mathbb{E}_{\sigma, x_\sigma \sim \mathcal{N}(x, \sigma)} [(x_\sigma - D(x_\sigma, \sigma))/\sigma] \frac{\partial g}{\partial \theta}.
\end{equation}
The gradient is not backpropagated through the denoiser as it is costly and unstable due to the noising step. In practice, only a single denoising step is applied for better performance. 


\subsection{Motivation}\label{sec:motiv}

As we can generate training functional maps easily (\cref{sec:FunctionalMapDiffusionModel}), our goal is to leverage the learned functional maps distribution from score models, for the matching of unseen shapes. 

A first solution could be to learn a conditional distribution by conditioning the functional map diffusion model on point descriptors to generate the target map, as in recent diffusion-based rigid shape-matching approaches~\cite{jiang2023se}. However, the learned model would be category-specific, as with classic deep functional maps methods.

A second solution could be to use the learned probability likelihood of the score model~\cite{song2021scorebased} as a proxy for measuring map quality. A similar idea, based on axiomatic constraints instead of learned penalties, has been explored in MapTree~\cite{ren2020maptree}. However, such an approach outputs a set of {\it candidate} maps, including symmetry-swapping maps, as purely intrinsic approaches do not differentiate between them. Moreover, evaluating the likelihood of a diffusion model is costly due to the integration of the generation trajectory.

A third potential solution is to adapt Score Distillation Sampling to the deep functional maps setting. Indeed, the recent Shape-Non-Rigid Kinematics (SNK)~\cite{attaiki2023shape} work has shown that deep functional maps networks can be used in a zero-shot setting. In this paper, the authors exploit Neural Correspondence Priors~\cite{attaiki2022ncp} to obtain good initializations and optimize the map using spectral and spatial axiomatic constraints.

We now discuss the adaptation of SDS to our problem.
Given two shapes $\mathcal{S}_1, \mathcal{S}_2$, the \textit{source domain} is shape correspondences, represented as functional maps (\cref{sec:funcmaps}). The \textit{target domain} is descriptor functions. The pointwise descriptors of shapes $F_i$ are parameterized by a deep neural network $F_i = f_\theta(\mathcal{S}_i)$, from which we estimate the functional map $C_{12}$ with the FMReg layer~\cite{donati2020deep}. In the SDS notation, $x$ is the functional map $C$, $\theta$ corresponds to the weights of the feature extractor, and $g$ is the FMReg layer. 

\begin{figure}[t]
    \centering
    \includegraphics[width=0.7\linewidth]{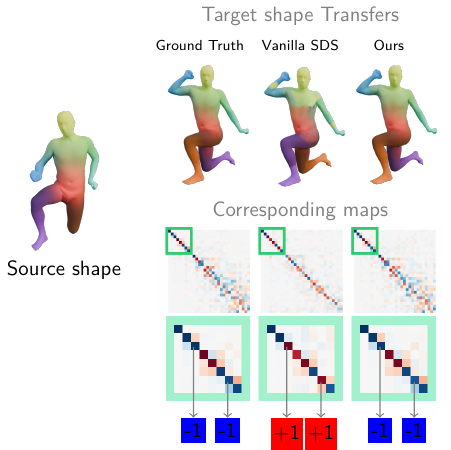}
    \caption{Comparison between vanilla SDS and our approach. The same diagonal structure is encouraged between the two approaches. However, using our proposed mask allows for fixing misalignments (sign flips in the functional maps, highlighted at the bottom) of the initialization. In contrast, the vanilla SDS converges towards the closest map with the same diagonal.}
    \label{fig:vanilla_nul}
    \vspace{-9pt}
\end{figure}

In~\cref{fig:vanilla_nul}, we perform a preliminary experiment with SDS. Notably, the recovered map shows significant mismatches (left and right legs are reversed) compared to the ground truth. 
We also observe that the structure promoted by the approach is nearly diagonal, as commonly observed in functional maps. The wrong signs along the diagonal cause the observed inconsistencies. We argue that the sign ambiguity of functional maps affects the performance (more discussions in the first section of supp. material). To better capture the underlying structure of the functional maps from data, we adopt a \textbf{sign-agnostic} approach in the following section.
\section{Method}
\label{sec:Method}

We first train a spectral diffusion model of functional maps (\cref{sec:FunctionalMapDiffusionModel}). Second, we devise a zero-shot training approach by distilling the knowledge learned by the trained model (\cref{sec:ProperSDS}), and the pipeline is illustrated in~\cref{fig:overview}. 

\subsection{Training}
\label{sec:FunctionalMapDiffusionModel}

Given a dataset of registered shapes, our objective is to train a spectral functional maps diffusion model $D(C_\sigma, \sigma)$. Using the registered shapes, we first build a dataset of ground-truth functional maps $C_{\text{gt}}$. Given $\mathcal{S}_1, \mathcal{S}_2$ two registered shapes, and $\Phi_1, \Phi_2$ their respective eigenfunctions, the ground-truth map between the two shapes is given by:
$$
C_{12-\text{gt}} = \Phi_2^ \dagger \Phi_1,
$$
where $(\cdot)^\dagger$ is the Moore-Penrose pseudoinverse.
We extract functional maps of fixed size $n \times n$ of template to shape correspondences. 

The extracted functional maps are thus matrices $C \in \mathcal{M}_n(\mathbb{R})$, which are analogous to images. We thus build upon the available image-based architectures and use the Diffusion Transformer architecture~\cite{peebles2023scalable}, which has shown great capabilities for image generation.

Our spectral denoiser $D_\psi(C_\sigma, \sigma)$ takes as input a matrix $C_\sigma \in \mathcal{M}_n(\mathbb{R})$ and noise level $\sigma$. To be sign-agnostic, we train a diffusion model on \textbf{absolute functional maps}, with input training data as the set of $|C_{\text{gt}}|$ (which is $C_{\text{gt}}$ with $x \to |x|$ applied on each element).


\subsection{Zero-shot shape matching}\label{sec:masku}

In this section, we are now given as input two {\it unseen shapes} $\mathcal{S}_1, \mathcal{S}_2$. We aim to estimate the functional map $C_{12}$ between the two shapes. We use the deep functional maps framework~\cite{donati2020deep} to differentiably estimate the functional map $C_{12}$, from pointwise descriptors $F_i = f_\theta(\mathcal{S}_i)$, parameterized by neural network weights $\theta$.

We optimize the parameters $\theta$ by applying a distillation loss to the functional maps. The remaining section discusses the construction of our distillation loss.  \newline

\par \noindent \textbf{Mask regularization} is a sign-agnostic regularization that has proven essential in deep functional maps as it provides reliable initialization towards the final solution~\cite{attaiki2022ncp}. In this section, we seek to provide a masked regularization in the form of~\cref{eq:fmap_compute}. We search for sparsity-promoting masks $M_\sigma$, such that given a ground truth map $C_{gt}$, we have:
\begin{equation}
||M_\sigma \cdot C_{gt}|| \simeq 0
\end{equation}
It is equivalent to say that $C_{gt}$ maximizes the likelihood 

\begin{equation}
p(C_\sigma; \sigma) \propto \text{exp}(-||M_\sigma \cdot C_\sigma||^2).
\end{equation}
Under this hypothesis, the score function derives as:

\begin{equation}
s(C_\sigma; \sigma) = \nabla_x \text{log}p(x:\sigma) = -2 M_\sigma^2 \cdot C_\sigma .
\end{equation}
By using~\Cref{eq:score}, we obtain:
\begin{equation}
M_\sigma^2 \cdot C_\sigma = ( C_\sigma - D(C_\sigma; \sigma))/2\sigma^2,
\end{equation}
which reduces as the following formula for computing $M_\sigma$ (by taking the mean over the noise distribution):
\begin{equation}
M_\sigma^2 = \mathbb{E}_{n_\sigma \sim \mathcal{N}(0, \sigma^2 I)}\left[(C_\sigma - D(C_\sigma; \sigma))/(2\sigma^2  C_\sigma) \right].
\end{equation}

Applying the formula directly would cause numerical instabilities when dividing by $C_\sigma$ that can contain 0 values, if arbitrary values of noise are sampled. We avoid this by sampling only $n_\sigma > 0$, which ensures only positive values when working with absolute functional maps $|C|$. The formula for the mask is finally:
\begin{equation}
\resizebox{0.85\linewidth}{!}{$
    M_\sigma^2 = \mathbb{E}_{n_\sigma \sim \mathcal{N}(0, \sigma^2 I), n_\sigma >0}\left[(|C|_\sigma - D(|C|_\sigma; \sigma))/(2\sigma^2  |C|_\sigma) \right].
    $}
\end{equation}

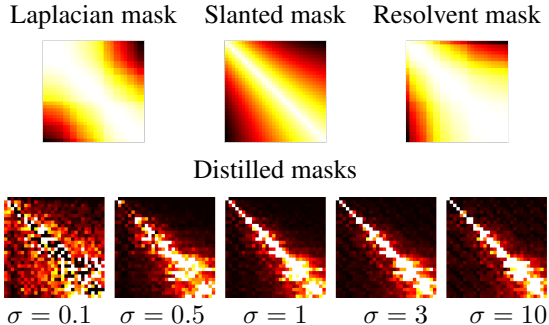
\begin{figure}
    \centering
    \begin{tikzpicture}
        
        \node (distilled) at (0,0) {
                        \includegraphics[height=0.16\linewidth]{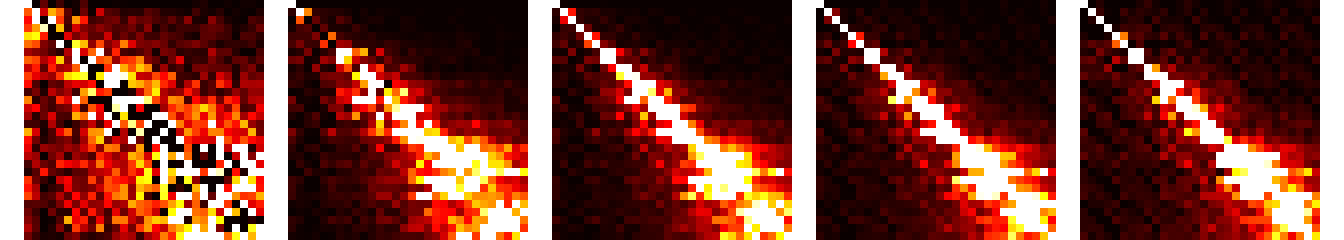}
                };
        \node [] at (-3., -.9)  {$\sigma = 0.1$};
        \node [] at (-1.5, -.9)  {$\sigma = 0.5$};
        \node [] at (0, -.9)  {$\sigma = 1$};
        \node [] at (1.6, -.9)  {$\sigma = 3$};
        \node [] at (3.1, -.9)  {$\sigma = 10$};
        \node [above=.em of distilled] (top) {Distilled masks};

        \node [draw, thick, lightgray, inner sep=0em, above=.3em of top] (slanted)  {
                        \includegraphics[height=0.16\linewidth]{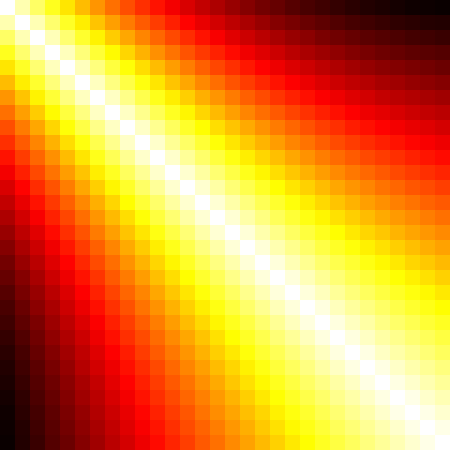}
                };
        \node [above=.3em of slanted]  {Slanted mask};
        \node [draw, thick, lightgray, inner sep=0em, left=3em of slanted] (orig) {
                        \includegraphics[height=0.16\linewidth]{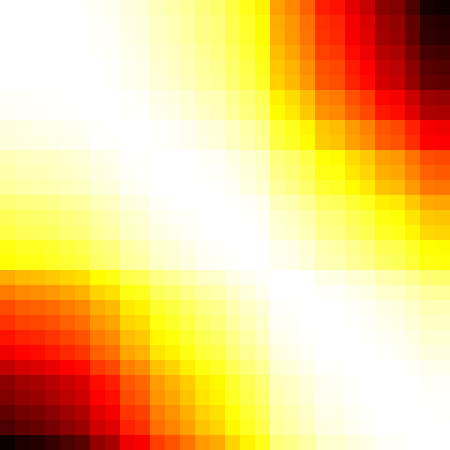}
                };
        \node [above=.1em of orig] {Laplacian mask};
        
        \node [draw, thick, lightgray, inner sep=0em, right=3em of slanted] (resolvent) {
                        \includegraphics[height=0.16\linewidth]{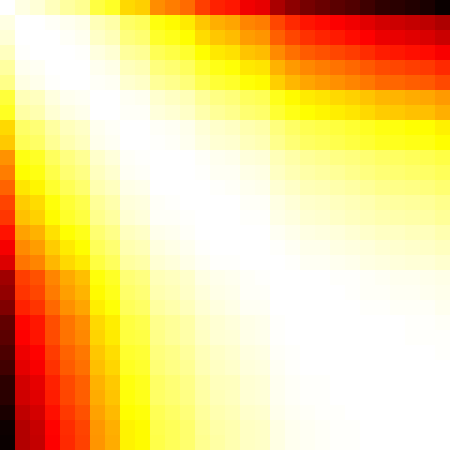}
                };
        \node [above=.3em of resolvent]  {Resolvent mask};
    \end{tikzpicture}
    \caption{Top: usual masks for functional map regularization. Bottom: estimated distillation masks at different noise levels}
    \label{fig:masks}
    \vspace{-9pt}
\end{figure}

We can distill the learned structure from the spectral diffusion model into a mask $M_\sigma$ for different noise levels, given any functional map matrix $C_{\text{init}}$. We show in~\cref{fig:masks} the estimated masks for different noise levels.
We can incorporate this mask into the functional map computation:
(1) we estimate a ``raw'' functional map $C_{\text{raw}}$ based on the input descriptors using \cref{eq:fmap_compute} with $\alpha = 0$;
(2) we then estimate a mask $M_\sigma$ from $C_{\text{raw}}$ and solve~\cref{eq:fmap_compute} with $M_\sigma$ to obtain a mask-regularized map $C_\text{reg}$. \newline

\begin{table*}[t]
    \centering
    \setlength\tabcolsep{6pt} 
    \begin{tabular}{c}
        \begin{adjustbox}{max width=\textwidth}
            \aboverulesep=0ex
            \belowrulesep=0ex
            \renewcommand{\arraystretch}{1.0}
            \begin{tabular}[t]{ccccccccc}
            \multicolumn{1}{c}{} & \multicolumn{1}{c}{} & 
            \multicolumn{3}{c}{Humans} & 
            \multicolumn{2}{c}{Humanoids} &
            \multicolumn{2}{c}{Animals}\\
            \vspace{-3.8ex} \\
            \multicolumn{1}{c}{} & \multicolumn{1}{c}{} & 
            \multicolumn{3}{c}{\downbracefill} & 
            \multicolumn{2}{c}{\downbracefill} &
            \multicolumn{2}{c}{\downbracefill}\\
            & Methods & FAUST & SCAPE & SHREC19 & DT4D-Intra & DT4D-Inter & SMAL & TOSCA \\
            \cmidrule(r){2-9}
            \vspace{-0.8ex} \\[-1.5ex]  
            \cmidrule(r){2-9}
            \ldelim\{{3}{.5cm}[{\rotatebox[origin=c]{90}{\small \textit{Axiom.}}}]
            & Ini + Zoomout (Laplacian) & 3.8 & 7.5 & 13.1 & 1.8 & 16.5 & 18.3 & 8.1 \\
            & Ini + Zoomout (Resolvent) & 3.2 & 5.7 & 12.4 & \textbf{1.6} & 13.4 & 19.1 & 5.4\\
            & Smooth shells~\cite{eisenberger2020smooth}           & 2.5 & 4.7 & 12.2 & / & / & 16.3 & / \\
            \addlinespace
            \ldelim\{{3}{.5cm}[{\rotatebox[origin=c]{90}{\small \textit{Learned}}}]
            & 3D-CODED~\cite{groueix2018_3DCODED}                 & 7.5 & 17.2 & 13.4 & 45.0 & 61.4 & 54.6 & 32.8\\
            & Neural Jacobian Fields~\cite{NJF_2022}    & 5.9 & 11.7 & 9.6 & 43.4 & 32.8 & 49.2 & 50.2 \\
            
            & Simplified Fmaps~\cite{magnet2024memory}           & \textbf{1.7} & \textbf{2.3} & \textbf{3.4} & 2.0 & \underline{8.9} & 42.1 & 5.1 \\
            \addlinespace
            \ldelim\{{2}{.5cm}[{\rotatebox[origin=c]{90}{\small \textit{Zero-shot}}}]
            & SNK~\cite{attaiki2023shape}                       & \underline{1.8} & 4.7 & 5.8 & 2.0 & 9.0 & \textbf{9.1} & \underline{3.6} \\
            \cmidrule(r){2-9}
            & Ini + Zoomout (our mask)  & 2.4 & 6.6 & 8.3 & 2.1 & 11.7 & 12.9 & 8.3 \\
            & Ours                      & 1.9 & \underline{4.4} & \underline{3.9} & \underline{1.8} & \textbf{8.6} & \underline{10.1} & \textbf{2.9} 
        \end{tabular} 
        \end{adjustbox}
    \end{tabular}
    \vspace{-3pt}
    \caption{Comparison of matching accuracy of axiomatic, learned, and zero-shot shape matching methods. The learning-based methods are trained on human shapes from Dynamic FAUST. The lower the better.\vspace{10pt}}
    \label{tab:big_table}
    \vspace{-18pt}
\end{table*}

\par \noindent \textbf{Proper SDS}\label{sec:ProperSDS} Similarly to SDS, we do not backpropagate through the mask optimization during optimization. Moreover, it has been shown that projecting the functional map on the ``proper" map space~\cite{ren2021discrete, attaiki2023understanding} (space of maps computed from a pointwise map) is necessary for convergence. We apply Zoomout \cite{zoomout} to the regularized functional map $C_\text{reg}$ and obtain a proper regularized map $C_{\text{proper}}$. We minimize the $L_2$ distance between the raw map and the proper map:
\begin{equation}
\mathcal{L}_{\text{proper}}(C_{\text{raw}}) = ||C_{\text{raw}} - C_{\text{proper}}||^2,
\end{equation}
where we only backpropagate to the feature extractor weights through $C_{raw}$. 
Finally, we also apply SDS to the absolute raw map. Thus, our total loss is :
\begin{equation}
    \mathcal{L}_{\text{total}}(C_{\text{raw}}) = \mathcal{L}_{\text{proper}}(C_{\text{raw}}) + \mathcal{L}_{SDS}(|C_{\text{raw}}|)
\label{eq:total}
\end{equation}

\myparagraph{Summary} Given a pair of shapes, and a trained spectral diffusion model, at test time we optimize a shape pointwise feature extractor, $F_i = f_\theta (\mathcal{S}_i)$. (1) Given features on the two shapes, we first estimate a functional map $C^{12}_{\text{raw}}$ between the shapes by solving~\cref{eq:fmap_compute} with $\alpha=0$. (2) We use this map to distill a mask $M_\sigma$ from the diffusion model, and solve~\cref{eq:fmap_compute} a second time to obtain a regularized map $C^{12}_{\text{reg}}$, on which we apply Zoomout to obtain $C^{12}_\text{proper}$. (3) This map, along with the diffusion model, is used to compute $\mathcal{L}_{\text{total}}$  (\cref{eq:total}). (4) The parameters $f_\theta$ of the feature extractor are optimized through back-propagation. (5) We convert the optimized functional map $C^{12}$ to a point-to-point map with the standard approach~\cite{funcmaps}. Note that our pipeline differs from previous deep functional maps approaches in that we \textit{avoid axiomatic priors}, such as Laplacian commutativity or orthogonality, both at mask estimation and training. Instead, all of our regularization and objective terms, except for the basic properness term, are derived \textit{solely from available training data}.

\section{Experiments}
\label{sec:Experiments}

\begin{figure*}
    \centering
    \includegraphics[width=0.8\linewidth]{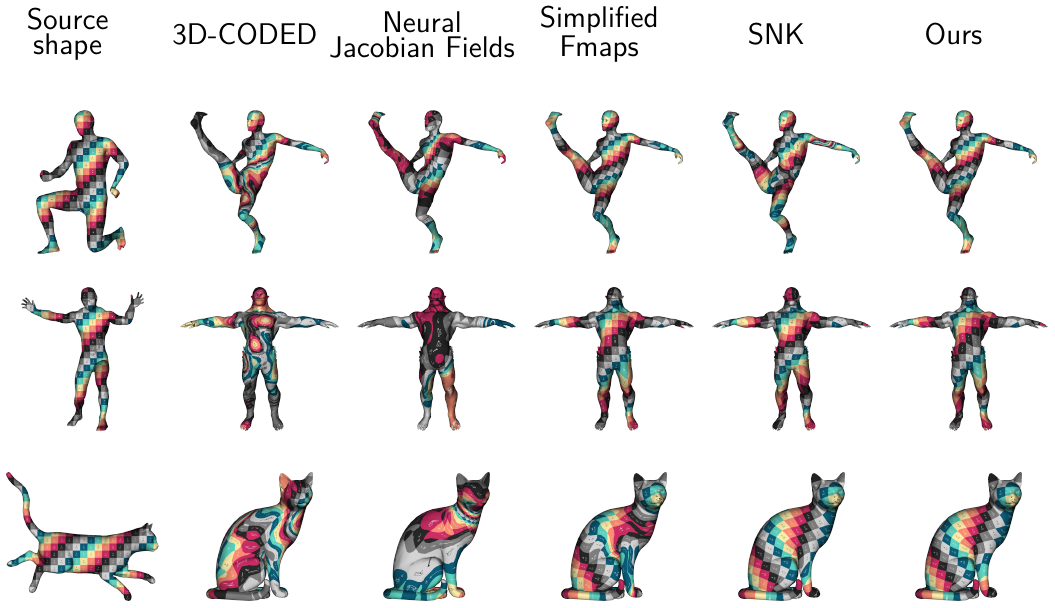}
    \caption{Texture transfer on different examples. We observe that deformation-based models such as 3D-CODED or Neural Jacobian Fields struggle with challenging poses and fail to generalize to unseen meshes. In the meantime, the learned pointwise descriptors of Simplified Fmaps generalize well to humanoids but struggle when confronted with new categories like animals. The SNK zero-shot approach can match shapes from different categories but struggles with challenging poses. Our approach can infer qualitative shape correspondences on each of those challenging examples.}
    \label{fig:texture_transfer}
\end{figure*}

\subsection{Experimental Details}
\myparagraph{Functional Map Diffusion Model.} 
We train our diffusion model on $30\times30$ maps, using template-to-shape maps on the D-FAUST dataset, for a total of  $\sim 40,000$ maps. The architecture is the DiT-S~\cite{peebles2023scalable} Diffusion Transformer with a patch size of 5. We train our model with EDM~\cite{karras2022elucidating} for 1000 epochs with the variance preserving loss.

\myparagraph{Zero-shot Optimization.} We use DiffusionNet~\cite{sharp2022diffusionnet} as our feature extractor. The estimated functional map size is $30\times30$. We follow the zero-shot experimental settings from SNK~\cite{attaiki2023shape}. We set $\sigma=1$ for our distilled mask as we found the best results from this specific value, with $N=100$ noisy samples, which can be done in a single batch denoising. Iterating the process did not provide significant improvement. We apply Zoomout to increase the map size from $30\times30$ to $40\times40$ to compute $\mathcal{L}_{proper}$.

\subsection{Datasets and Comparison}

\myparagraph{Near-isometric Shape Matching.} We first test the generalization of our approach on human data. Notably, we test on the oriented versions of the remeshed FAUST~\cite{bogo2014faust}, SCAPE~\cite{anguelov2005scape}, and SHREC~\cite{melzi2019shrec} datasets, on the usual test sets from commonly used train/test splits. 

\myparagraph{Non-isometric Shape Matching.} We then test our approach on unseen data types, namely humanoids and animals. For humanoids, we used the remeshed split of the DynamicThings4D dataset (DT4D)~\cite{li20214dcomplete}, from which we use the intra-category and inter-category test sets from~\cite{li2022learning}. For animals, we tested our approach on the SMAL remeshed dataset~\cite{donati2022complex} and animal shape pairs from the TOSCA dataset~\cite{bronstein2008numerical}, as done in~\cite{sun2023spatially}. 

\myparagraph{Baselines.} We compare our method with axiomatic~\cite{eisenberger2020smooth}, learned~\cite{NJF_2022, groueix2018_3DCODED, magnet2024memory} and zero-shot~\cite{attaiki2023shape} baselines. A detailed description is provided in the supplementary material.

\subsection{Results}

\change{We follow the Princeton benchmark evaluation protocol~\cite{kim2011blended} and evaluate the accuracy of the maps using the geodesic error of the computed correspondence.} We present the results on near isometric data on the left of~\cref{tab:big_table}. We outperform both axiomatic and other zero-shot approaches on this task. Notably, we are close to the state-of-the-art deep functional map Simplified Fmaps approach~\cite{magnet2024memory}. Also, our distilled mask provides, in general, a good quality initialization, competitive with other approaches, and outperforms the Laplacian and Resolvent masks (Ini + Zoomout).
\par 
The results on non-isometric data are in the right of~\cref{tab:big_table}. Notably, our approach outperforms other approaches on the DT4D-Inter challenge and the TOSCA dataset, including Simplified Fmaps. 3D-CODED and Neural Jacobian Fields, based on learned deformation models, perform poorly on humanoids and animals, as the learned deformations are category-specific. Moreover, Simplified Fmaps fails on the SMAL dataset, suggesting that learned descriptors do not generalize well to animals. We also outperform SNK on most datasets, and our distilled mask provides a better initialization than the traditional Laplacian and resolvent masks (Ini + Zoomout).

\subsection{Ablation study}

We ablate the different components of our approach in~\Cref{tab:ablation}. As stated in~\cref{sec:motiv}, vanilla SDS fails to correct misalignments and shows poor results. Using $\mathcal{L}_{proper}$ alone is efficient but far from state-of-the-art performance. The best is to combine $\mathcal{L}_{proper}$ and $\mathcal{L}_{SDS}$. Finally, we added axiomatic penalties (orthogonal, bijectivity, and Laplacian losses) to DiffuMatch. We find that the final accuracy is nearly the same, indicating that our formulation already encompasses these axiomatic regularizations.
\begin{table}[!h]
\begin{tabular}{lc}
Approach        & Geod Error (SHREC) \\ \hline\hline
Vanilla SDS & 57.3  \\
Mask + Zoomout & 8.3 \\
$\mathcal{L}_{proper}$ & 7.7 \\
Mask + $\mathcal{L}_{SDS}$ & 7.1 \\
Mask + $\mathcal{L}_{proper}$ & 6.7 \\\hline
Mask + $\mathcal{L}_{proper}$ + $\mathcal{L}_{SDS}$ (ours) & {\bf 4.4} \\
Ours + Axiomatic & 4.3
\end{tabular}
\vspace{-3pt}
\caption{Ablation study of the different components of our approach}\label{tab:ablation}
\vspace{-18pt}
\end{table}
\section{Limitations and future work}
Despite the efficiency of our method, our method might not handle well highly non-isometric shapes or partial shapes~\cite{rodola2017partial}, which is a known issue of functional map-based methods~\cite{bracha2024unsupervised, bracha2025partial}. A potential direction to mitigate this problem is to jointly learn the basis along with spectral regularization.
Moreover, our diffusion model is trained only on human shapes with limited diversity. Using or generating more registered training data will be crucial towards a unified model for functional maps.

\section{Conclusion}
In this work, we presented a functional map score generative model, trained on registered human shapes, to learn the structure induced by the distribution of functional maps. Based on our model, we proposed a novel functional map penalty and a zero-shot training pipeline to match shape pairs at test time. The results demonstrate state-of-the-art results for zero-shot shape matching on diverse benchmarks, including categories unseen during training. We believe our approach will serve as a first step towards foundation models for shape matching.
\label{sec:Conclusion}

\paragraph{Acknowledgements} This work was performed using HPC resources from GENCI-IDRIS (Grant 2025-AD010613760R2). Parts of this work were supported by the ERC Consolidator Grant 101087347 (VEGA), as well as gifts from Ansys and Adobe Research, and the ERC Starting Grant SpatialSem (101076253).

{
    \small
    \bibliographystyle{ieeenat_fullname}
    \bibliography{main}

\begin{thebibliography}{84}
\providecommand{\natexlab}[1]{#1}
\providecommand{\url}[1]{\texttt{#1}}
\expandafter\ifx\csname urlstyle\endcsname\relax
  \providecommand{\doi}[1]{doi: #1}\else
  \providecommand{\doi}{doi: \begingroup \urlstyle{rm}\Url}\fi

\bibitem[Aigerman et~al.(2022)Aigerman, Gupta, Kim, Chaudhuri, Saito, and Groueix]{NJF_2022}
Noam Aigerman, Kunal Gupta, Vladimir~G. Kim, Siddhartha Chaudhuri, Jun Saito, and Thibault Groueix.
\newblock Neural jacobian fields: learning intrinsic mappings of arbitrary meshes.
\newblock \emph{ACM Trans. Graph.}, 41\penalty0 (4), 2022.

\bibitem[Alexa et~al.(2000)Alexa, Cohen-Or, and Levin]{alexa2023rigid}
Marc Alexa, Daniel Cohen-Or, and David Levin.
\newblock As-rigid-as-possible shape interpolation.
\newblock In \emph{Proceedings of the 27th Annual Conference on Computer Graphics and Interactive Techniques}, page 157–164, USA, 2000. ACM Press/Addison-Wesley Publishing Co.

\bibitem[Anguelov et~al.(2023)Anguelov, Srinivasan, Koller, Thrun, Rodgers, and Davis]{anguelov2005scape}
Dragomir Anguelov, Praveen Srinivasan, Daphne Koller, Sebastian Thrun, Jim Rodgers, and James Davis.
\newblock Scape: Shape completion and animation of people.
\newblock 2023.

\bibitem[Attaiki and Ovsjanikov(2022)]{attaiki2022ncp}
Souhaib Attaiki and Maks Ovsjanikov.
\newblock Ncp: Neural correspondence prior for effective unsupervised shape matching.
\newblock \emph{Advances in Neural Information Processing Systems}, 35:\penalty0 28842--28857, 2022.

\bibitem[Attaiki and Ovsjanikov(2023{\natexlab{a}})]{attaiki2023shape}
Souhaib Attaiki and Maks Ovsjanikov.
\newblock Shape non-rigid kinematics (snk): A zero-shot method for non-rigid shape matching via unsupervised functional map regularized reconstruction.
\newblock \emph{Advances in Neural Information Processing Systems}, 36:\penalty0 70012--70032, 2023{\natexlab{a}}.

\bibitem[Attaiki and Ovsjanikov(2023{\natexlab{b}})]{attaiki2023understanding}
Souhaib Attaiki and Maks Ovsjanikov.
\newblock Understanding and improving features learned in deep functional maps.
\newblock In \emph{Proceedings of the IEEE/CVF Conference on Computer Vision and Pattern Recognition}, pages 1316--1326, 2023{\natexlab{b}}.

\bibitem[Bensa{\"\i}d et~al.(2023)Bensa{\"\i}d, Bracha, and Kimmel]{bensaid2023partial}
David Bensa{\"\i}d, Amit Bracha, and Ron Kimmel.
\newblock Partial shape similarity by multi-metric hamiltonian spectra matching.
\newblock In \emph{International Conference on Scale Space and Variational Methods in Computer Vision}, pages 717--729. Springer, 2023.

\bibitem[Bogo et~al.(2014)Bogo, Romero, Loper, and Black]{bogo2014faust}
Federica Bogo, Javier Romero, Matthew Loper, and Michael~J Black.
\newblock Faust: Dataset and evaluation for 3d mesh registration.
\newblock In \emph{Proceedings of the IEEE conference on computer vision and pattern recognition}, pages 3794--3801, 2014.

\bibitem[Bogo et~al.(2017)Bogo, Romero, Pons-Moll, and Black]{bogo2017dynamic}
Federica Bogo, Javier Romero, Gerard Pons-Moll, and Michael~J Black.
\newblock Dynamic faust: Registering human bodies in motion.
\newblock In \emph{Proceedings of the IEEE conference on computer vision and pattern recognition}, pages 6233--6242, 2017.

\bibitem[Bracha et~al.(2020)Bracha, Halim, and Kimmel]{bracha2020corres}
Amit Bracha, Oshri Halim, and Ron Kimmel.
\newblock {Shape Correspondence by Aligning Scale-invariant LBO Eigenfunctions}.
\newblock In \emph{Eurographics Workshop on 3D Object Retrieval}. The Eurographics Association, 2020.

\bibitem[Bracha et~al.(2024)Bracha, Dag{\`e}s, and Kimmel]{bracha2024unsupervised}
Amit Bracha, Thomas Dag{\`e}s, and Ron Kimmel.
\newblock On unsupervised partial shape correspondence.
\newblock In \emph{Proceedings of the Asian Conference on Computer Vision}, pages 4488--4504, 2024.

\bibitem[Bracha et~al.(2025)Bracha, Dag\`{e}s, and Kimmel]{bracha2025partial}
Amit Bracha, Thomas Dag\`{e}s, and Ron Kimmel.
\newblock Wormhole loss for partial shape matching.
\newblock In \emph{Proceedings of the 38th International Conference on Neural Information Processing Systems}, 2025.

\bibitem[Bronstein et~al.(2008)Bronstein, Bronstein, and Kimmel]{bronstein2008numerical}
Alexander~M Bronstein, Michael~M Bronstein, and Ron Kimmel.
\newblock \emph{Numerical geometry of non-rigid shapes}.
\newblock Springer Science \& Business Media, 2008.

\bibitem[Cao and Bernard(2022)]{cao2022unsupervised}
Dongliang Cao and Florian Bernard.
\newblock Unsupervised deep multi-shape matching.
\newblock In \emph{European Conference on Computer Vision}, pages 55--71. Springer, 2022.

\bibitem[Cao et~al.(2023)Cao, Roetzer, and Bernard]{Cao2023UnsupervisedLO}
Dongliang Cao, Paul Roetzer, and Florian Bernard.
\newblock Unsupervised learning of robust spectral shape matching.
\newblock \emph{ACM Transactions on Graphics (TOG)}, 42:\penalty0 1 -- 15, 2023.

\bibitem[Cao et~al.(2024{\natexlab{a}})Cao, Eisenberger, El~Amrani, Cremers, and Bernard]{cao2024spectral}
Dongliang Cao, Marvin Eisenberger, Nafie El~Amrani, Daniel Cremers, and Florian Bernard.
\newblock Spectral meets spatial: Harmonising 3d shape matching and interpolation.
\newblock In \emph{Proceedings of the IEEE/CVF Conference on Computer Vision and Pattern Recognition}, pages 3658--3668, 2024{\natexlab{a}}.

\bibitem[Cao et~al.(2024{\natexlab{b}})Cao, L{\"a}hner, and Bernard]{cao2024synchronous}
Dongliang Cao, Zorah L{\"a}hner, and Florian Bernard.
\newblock Synchronous diffusion for unsupervised smooth non-rigid 3d shape matching.
\newblock In \emph{European Conference on Computer Vision}, pages 262--281. Springer, 2024{\natexlab{b}}.

\bibitem[Dhariwal and Nichol(2021)]{dhariwal2021diffusion}
Prafulla Dhariwal and Alexander Nichol.
\newblock Diffusion models beat gans on image synthesis.
\newblock \emph{Advances in neural information processing systems}, 34:\penalty0 8780--8794, 2021.

\bibitem[Dinh et~al.(2005)Dinh, Yezzi, and Turk]{dinh2005texture}
Huong~Quynh Dinh, Anthony Yezzi, and Greg Turk.
\newblock Texture transfer during shape transformation.
\newblock \emph{ACM Transactions on Graphics (ToG)}, 24\penalty0 (2):\penalty0 289--310, 2005.

\bibitem[Donati et~al.(2020)Donati, Sharma, and Ovsjanikov]{donati2020deep}
Nicolas Donati, Abhishek Sharma, and Maks Ovsjanikov.
\newblock Deep geometric functional maps: Robust feature learning for shape correspondence.
\newblock In \emph{Proceedings of the IEEE/CVF Conference on Computer Vision and Pattern Recognition}, pages 8592--8601, 2020.

\bibitem[Donati et~al.(2022)Donati, Corman, Melzi, and Ovsjanikov]{donati2022complex}
Nicolas Donati, Etienne Corman, Simone Melzi, and Maks Ovsjanikov.
\newblock Complex functional maps: A conformal link between tangent bundles.
\newblock In \emph{Computer Graphics Forum}, pages 317--334. Wiley Online Library, 2022.

\bibitem[Ehm et~al.(2024{\natexlab{a}})Ehm, Gao, Roetzer, Eisenberger, Cremers, and Bernard]{ehm2024partial}
Viktoria Ehm, Maolin Gao, Paul Roetzer, Marvin Eisenberger, Daniel Cremers, and Florian Bernard.
\newblock Partial-to-partial shape matching with geometric consistency.
\newblock In \emph{Proceedings of the IEEE/CVF Conference on Computer Vision and Pattern Recognition}, pages 27488--27497, 2024{\natexlab{a}}.

\bibitem[Ehm et~al.(2024{\natexlab{b}})Ehm, Roetzer, Eisenberger, Gao, Bernard, and Cremers]{ehm2024geometrically}
Viktoria Ehm, Paul Roetzer, Marvin Eisenberger, Maolin Gao, Florian Bernard, and Daniel Cremers.
\newblock Geometrically consistent partial shape matching.
\newblock In \emph{2024 International Conference on 3D Vision (3DV)}, pages 914--922. IEEE, 2024{\natexlab{b}}.

\bibitem[Eisenberger et~al.(2020{\natexlab{a}})Eisenberger, Lahner, and Cremers]{eisenberger2020smooth}
Marvin Eisenberger, Zorah Lahner, and Daniel Cremers.
\newblock Smooth shells: Multi-scale shape registration with functional maps.
\newblock In \emph{Proceedings of the IEEE/CVF Conference on Computer Vision and Pattern Recognition}, pages 12265--12274, 2020{\natexlab{a}}.

\bibitem[Eisenberger et~al.(2020{\natexlab{b}})Eisenberger, Toker, Leal-Taix{\'e}, and Cremers]{eisenberger2020deep}
Marvin Eisenberger, Aysim Toker, Laura Leal-Taix{\'e}, and Daniel Cremers.
\newblock Deep shells: Unsupervised shape correspondence with optimal transport.
\newblock \emph{Advances in Neural information processing systems}, 33:\penalty0 10491--10502, 2020{\natexlab{b}}.

\bibitem[Eisenberger et~al.(2023)Eisenberger, Toker, Leal-Taix{\'e}, and Cremers]{eisenberger2023g}
Marvin Eisenberger, Aysim Toker, Laura Leal-Taix{\'e}, and Daniel Cremers.
\newblock G-msm: Unsupervised multi-shape matching with graph-based affinity priors.
\newblock \emph{Proceedings of the IEEE/CVF Conference on Computer Vision and Pattern Recognition}, 2023.

\bibitem[Fumero et~al.()Fumero, Pegoraro, Maiorca, Locatello, and Rodol{\`a}]{fumerolatent}
Marco Fumero, Marco Pegoraro, Valentino Maiorca, Francesco Locatello, and Emanuele Rodol{\`a}.
\newblock Latent functional maps: a spectral framework for representation alignment.
\newblock In \emph{The Thirty-eighth Annual Conference on Neural Information Processing Systems}.

\bibitem[Gao et~al.(2021)Gao, Lahner, Thunberg, Cremers, and Bernard]{gao2021isometric}
Maolin Gao, Zorah Lahner, Johan Thunberg, Daniel Cremers, and Florian Bernard.
\newblock Isometric multi-shape matching.
\newblock In \emph{Proceedings of the IEEE/CVF Conference on Computer Vision and Pattern Recognition}, pages 14183--14193, 2021.

\bibitem[Gao et~al.(2023)Gao, Roetzer, Eisenberger, L{\"a}hner, Moeller, Cremers, and Bernard]{gao2023sigma}
Maolin Gao, Paul Roetzer, Marvin Eisenberger, Zorah L{\"a}hner, Michael Moeller, Daniel Cremers, and Florian Bernard.
\newblock Sigma: Scale-invariant global sparse shape matching.
\newblock In \emph{Proceedings of the IEEE/CVF International Conference on Computer Vision}, pages 645--654, 2023.

\bibitem[Groueix et~al.(2018)Groueix, Fisher, Kim, Russell, and Aubry]{groueix2018_3DCODED}
Thibault Groueix, Matthew Fisher, Vladimir~G. Kim, Bryan Russell, and Mathieu Aubry.
\newblock 3d-coded : 3d correspondences by deep deformation.
\newblock In \emph{ECCV}, 2018.

\bibitem[Halimi and Kimmel(2018)]{halimi2018self}
Oshri Halimi and Ron Kimmel.
\newblock Self functional maps.
\newblock In \emph{2018 International Conference on 3D Vision (3DV)}, pages 710--718. IEEE, 2018.

\bibitem[Halimi et~al.(2019)Halimi, Litany, Rodola, Bronstein, and Kimmel]{halimi2019unsupervised}
Oshri Halimi, Or Litany, Emanuele Rodola, Alex~M Bronstein, and Ron Kimmel.
\newblock Unsupervised learning of dense shape correspondence.
\newblock In \emph{Proceedings of the IEEE/CVF Conference on Computer Vision and Pattern Recognition}, pages 4370--4379, 2019.

\bibitem[Ho et~al.(2020)Ho, Jain, and Abbeel]{ho2020denoising}
Jonathan Ho, Ajay Jain, and Pieter Abbeel.
\newblock Denoising diffusion probabilistic models.
\newblock \emph{Advances in neural information processing systems}, 33:\penalty0 6840--6851, 2020.

\bibitem[Huang et~al.(2021)Huang, Huang, Sun, Zhang, Jiang, and Bajaj]{ARAPREG_2021}
Q. Huang, X. Huang, B. Sun, Z. Zhang, J. Jiang, and C. Bajaj.
\newblock Arapreg: An as-rigid-as possible regularization loss for learning deformable shape generators.
\newblock In \emph{2021 IEEE/CVF International Conference on Computer Vision (ICCV)}, pages 5795--5805, Los Alamitos, CA, USA, 2021. IEEE Computer Society.

\bibitem[Hyv{\"a}rinen and Dayan(2005)]{hyvarinen2005estimation}
Aapo Hyv{\"a}rinen and Peter Dayan.
\newblock Estimation of non-normalized statistical models by score matching.
\newblock \emph{Journal of Machine Learning Research}, 6\penalty0 (4), 2005.

\bibitem[Jiang et~al.(2023)Jiang, Salzmann, Dang, Xie, and Yang]{jiang2023se}
Haobo Jiang, Mathieu Salzmann, Zheng Dang, Jin Xie, and Jian Yang.
\newblock Se (3) diffusion model-based point cloud registration for robust 6d object pose estimation.
\newblock \emph{Advances in Neural Information Processing Systems}, 36:\penalty0 21285--21297, 2023.

\bibitem[Karras et~al.(2022)Karras, Aittala, Aila, and Laine]{karras2022elucidating}
Tero Karras, Miika Aittala, Timo Aila, and Samuli Laine.
\newblock Elucidating the design space of diffusion-based generative models.
\newblock \emph{Advances in neural information processing systems}, 35:\penalty0 26565--26577, 2022.

\bibitem[Kim et~al.(2011)Kim, Lipman, and Funkhouser]{kim2011blended}
Vladimir~G Kim, Yaron Lipman, and Thomas Funkhouser.
\newblock Blended intrinsic maps.
\newblock \emph{ACM transactions on graphics (TOG)}, 30\penalty0 (4):\penalty0 1--12, 2011.

\bibitem[Li et~al.(2022{\natexlab{a}})Li, Attaiki, and Ovsjanikov]{li2022srfeat}
Lei Li, Souhaib Attaiki, and Maks Ovsjanikov.
\newblock Srfeat: Learning locally accurate and globally consistent non-rigid shape correspondence.
\newblock In \emph{2022 International Conference on 3D Vision (3DV)}, pages 144--154. IEEE, 2022{\natexlab{a}}.

\bibitem[Li et~al.(2022{\natexlab{b}})Li, Donati, and Ovsjanikov]{li2022learning}
Lei Li, Nicolas Donati, and Maks Ovsjanikov.
\newblock Learning multi-resolution functional maps with spectral attention for robust shape matching.
\newblock \emph{Advances in Neural Information Processing Systems}, 35:\penalty0 29336--29349, 2022{\natexlab{b}}.

\bibitem[Li et~al.(2021)Li, Takehara, Taketomi, Zheng, and Nie{\ss}ner]{li20214dcomplete}
Yang Li, Hikari Takehara, Takafumi Taketomi, Bo Zheng, and Matthias Nie{\ss}ner.
\newblock 4dcomplete: Non-rigid motion estimation beyond the observable surface.
\newblock In \emph{Proceedings of the IEEE/CVF International Conference on Computer Vision}, pages 12706--12716, 2021.

\bibitem[Litany et~al.(2017)Litany, Remez, Rodola, Bronstein, and Bronstein]{litany2017deep}
Or Litany, Tal Remez, Emanuele Rodola, Alex Bronstein, and Michael Bronstein.
\newblock Deep functional maps: Structured prediction for dense shape correspondence.
\newblock In \emph{Proceedings of the IEEE international conference on computer vision}, pages 5659--5667, 2017.

\bibitem[Lugmayr et~al.(2022)Lugmayr, Danelljan, Romero, Yu, Timofte, and Van~Gool]{lugmayr2022repaint}
Andreas Lugmayr, Martin Danelljan, Andres Romero, Fisher Yu, Radu Timofte, and Luc Van~Gool.
\newblock Repaint: Inpainting using denoising diffusion probabilistic models.
\newblock In \emph{Proceedings of the IEEE/CVF conference on computer vision and pattern recognition}, pages 11461--11471, 2022.

\bibitem[Lukoianov et~al.(2024)Lukoianov, de~Oc{\'a}riz~Borde, Greenewald, Guizilini, Bagautdinov, Sitzmann, and Solomon]{lukoianov2024score}
Artem Lukoianov, Haitz~S{\'a}ez de Oc{\'a}riz~Borde, Kristjan Greenewald, Vitor~Campagnolo Guizilini, Timur Bagautdinov, Vincent Sitzmann, and Justin Solomon.
\newblock Score distillation via reparametrized {DDIM}.
\newblock In \emph{The Thirty-eighth Annual Conference on Neural Information Processing Systems}, 2024.

\bibitem[Magnet and Ovsjanikov(2024)]{magnet2024memory}
Robin Magnet and Maks Ovsjanikov.
\newblock Memory-scalable and simplified functional map learning.
\newblock In \emph{Proceedings of the IEEE/CVF Conference on Computer Vision and Pattern Recognition}, pages 4041--4050, 2024.

\bibitem[Magnet et~al.(2022)Magnet, Ren, Sorkine-Hornung, and Ovsjanikov]{magnet2022smooth}
Robin Magnet, Jing Ren, Olga Sorkine-Hornung, and Maks Ovsjanikov.
\newblock Smooth non-rigid shape matching via effective dirichlet energy optimization.
\newblock In \emph{International Conference on 3D Vision (3DV)}, 2022.

\bibitem[Melzi et~al.(2019{\natexlab{a}})Melzi, Marin, Rodol{\`a}, Castellani, Ren, Poulenard, Ovsjanikov, et~al.]{melzi2019shrec}
Simone Melzi, Riccardo Marin, Emanuele Rodol{\`a}, Umberto Castellani, Jing Ren, Adrien Poulenard, P Ovsjanikov, et~al.
\newblock Shrec’19: matching humans with different connectivity.
\newblock In \emph{Eurographics Workshop on 3D Object Retrieval}, pages 1--8. The Eurographics Association, 2019{\natexlab{a}}.

\bibitem[Melzi et~al.(2019{\natexlab{b}})Melzi, Ren, Rodol\`{a}, Sharma, Wonka, and Ovsjanikov]{zoomout}
Simone Melzi, Jing Ren, Emanuele Rodol\`{a}, Abhishek Sharma, Peter Wonka, and Maks Ovsjanikov.
\newblock Zoomout: spectral upsampling for efficient shape correspondence.
\newblock \emph{ACM Trans. Graph.}, 38\penalty0 (6), 2019{\natexlab{b}}.

\bibitem[Meng et~al.(2022)Meng, He, Song, Song, Wu, Zhu, and Ermon]{meng2021sdedit}
Chenlin Meng, Yutong He, Yang Song, Jiaming Song, Jiajun Wu, Jun-Yan Zhu, and Stefano Ermon.
\newblock {SDE}dit: Guided image synthesis and editing with stochastic differential equations.
\newblock 2022.

\bibitem[Merrouche et~al.(2023)Merrouche, Regateiro, Wuhrer, and Boyer]{Merrouche_2023_BMVC}
Aymen Merrouche, Joao Pedro~Cova Regateiro, Stefanie Wuhrer, and Edmond Boyer.
\newblock Deformation-guided unsupervised non-rigid shape matching.
\newblock In \emph{34th British Machine Vision Conference 2023, {BMVC} 2023, Aberdeen, UK, November 20-24, 2023}. BMVA, 2023.

\bibitem[Ovsjanikov et~al.(2012)Ovsjanikov, Ben-Chen, Solomon, Butscher, and Guibas]{funcmaps}
Maks Ovsjanikov, Mirela Ben-Chen, Justin Solomon, Adrian Butscher, and Leonidas Guibas.
\newblock Functional maps: a flexible representation of maps between shapes.
\newblock \emph{ACM Trans. Graph.}, 31\penalty0 (4), 2012.

\bibitem[Ovsjanikov et~al.(2016)Ovsjanikov, Corman, Bronstein, Rodol\`{a}, Ben-Chen, Guibas, Chazal, and Bronstein]{ovsjanikov2016computing}
Maks Ovsjanikov, Etienne Corman, Michael Bronstein, Emanuele Rodol\`{a}, Mirela Ben-Chen, Leonidas Guibas, Frederic Chazal, and Alex Bronstein.
\newblock Computing and processing correspondences with functional maps.
\newblock In \emph{SIGGRAPH ASIA 2016 Courses}, New York, NY, USA, 2016. Association for Computing Machinery.

\bibitem[Pai et~al.(2021)Pai, Ren, Melzi, Wonka, and Ovsjanikov]{pai2021fast}
Gautam Pai, Jing Ren, Simone Melzi, Peter Wonka, and Maks Ovsjanikov.
\newblock Fast sinkhorn filters: Using matrix scaling for non-rigid shape correspondence with functional maps.
\newblock In \emph{Proceedings of the IEEE/CVF Conference on Computer Vision and Pattern Recognition}, pages 384--393, 2021.

\bibitem[Panine et~al.(2022)Panine, Kirgo, and Ovsjanikov]{panine2022non}
Mikhail Panine, Maxime Kirgo, and Maks Ovsjanikov.
\newblock Non-isometric shape matching via functional maps on landmark-adapted bases.
\newblock In \emph{Computer graphics forum}, pages 394--417. Wiley Online Library, 2022.

\bibitem[Parisi(1981)]{parisi1981correlation}
Giorgio Parisi.
\newblock Correlation functions and computer simulations.
\newblock \emph{Nuclear Physics B}, 180\penalty0 (3):\penalty0 378--384, 1981.

\bibitem[Peebles and Xie(2023)]{peebles2023scalable}
William Peebles and Saining Xie.
\newblock Scalable diffusion models with transformers.
\newblock In \emph{Proceedings of the IEEE/CVF international conference on computer vision}, pages 4195--4205, 2023.

\bibitem[Poole et~al.(2023)Poole, Jain, Barron, and Mildenhall]{poole2023dreamfusion}
Ben Poole, Ajay Jain, Jonathan~T. Barron, and Ben Mildenhall.
\newblock Dreamfusion: Text-to-3d using 2d diffusion.
\newblock In \emph{The Eleventh International Conference on Learning Representations}, 2023.

\bibitem[Ren et~al.(2018)Ren, Poulenard, Wonka, and Ovsjanikov]{ren2018continuous}
Jing Ren, Adrien Poulenard, Peter Wonka, and Maks Ovsjanikov.
\newblock Continuous and orientation-preserving correspondences via functional maps.
\newblock \emph{ACM Transactions on Graphics (ToG)}, 37\penalty0 (6):\penalty0 1--16, 2018.

\bibitem[Ren et~al.(2019)Ren, Panine, Wonka, and Ovsjanikov]{ren2019structured}
Jing Ren, Mikhail Panine, Peter Wonka, and Maks Ovsjanikov.
\newblock Structured regularization of functional map computations.
\newblock In \emph{Computer Graphics Forum}, pages 39--53. Wiley Online Library, 2019.

\bibitem[Ren et~al.(2020)Ren, Melzi, Ovsjanikov, and Wonka]{ren2020maptree}
Jing Ren, Simone Melzi, Maks Ovsjanikov, and Peter Wonka.
\newblock Maptree: Recovering multiple solutions in the space of maps.
\newblock \emph{ACM Transactions on Graphics}, 39\penalty0 (6):\penalty0 1--17, 2020.

\bibitem[Ren et~al.(2021)Ren, Melzi, Wonka, and Ovsjanikov]{ren2021discrete}
Jing Ren, Simone Melzi, Peter Wonka, and Maks Ovsjanikov.
\newblock Discrete optimization for shape matching.
\newblock In \emph{Computer Graphics Forum}, pages 81--96. Wiley Online Library, 2021.

\bibitem[Rodol{\`a} et~al.(2017)Rodol{\`a}, Cosmo, Bronstein, Torsello, and Cremers]{rodola2017partial}
Emanuele Rodol{\`a}, Luca Cosmo, Michael~M Bronstein, Andrea Torsello, and Daniel Cremers.
\newblock Partial functional correspondence.
\newblock In \emph{Computer graphics forum}, pages 222--236. Wiley Online Library, 2017.

\bibitem[Roetzer and Bernard(2024)]{roetzer2024spidermatch}
Paul Roetzer and Florian Bernard.
\newblock Spidermatch: 3d shape matching with global optimality and geometric consistency.
\newblock In \emph{Proceedings of the IEEE/CVF Conference on Computer Vision and Pattern Recognition}, pages 14543--14553, 2024.

\bibitem[Roetzer et~al.(2024)Roetzer, Abbas, Cao, Bernard, and Swoboda]{roetzer2024discomatch}
Paul Roetzer, Ahmed Abbas, Dongliang Cao, Florian Bernard, and Paul Swoboda.
\newblock Discomatch: Fast discrete optimisation for geometrically consistent 3d shape matching.
\newblock In \emph{European Conference on Computer Vision}, pages 443--460. Springer, 2024.

\bibitem[Rombach et~al.(2022)Rombach, Blattmann, Lorenz, Esser, and Ommer]{rombach2022high}
Robin Rombach, Andreas Blattmann, Dominik Lorenz, Patrick Esser, and Bj{\"o}rn Ommer.
\newblock High-resolution image synthesis with latent diffusion models.
\newblock In \emph{Proceedings of the IEEE/CVF conference on computer vision and pattern recognition}, pages 10684--10695, 2022.

\bibitem[Roufosse et~al.(2019)Roufosse, Sharma, and Ovsjanikov]{roufosse2019unsupervised}
Jean-Michel Roufosse, Abhishek Sharma, and Maks Ovsjanikov.
\newblock Unsupervised deep learning for structured shape matching.
\newblock In \emph{Proceedings of the IEEE/CVF International Conference on Computer Vision}, pages 1617--1627, 2019.

\bibitem[Sharma and Ovsjanikov(2020)]{sharma2020weakly}
Abhishek Sharma and Maks Ovsjanikov.
\newblock Weakly supervised deep functional maps for shape matching.
\newblock \emph{Advances in Neural Information Processing Systems}, 33:\penalty0 19264--19275, 2020.

\bibitem[Sharp et~al.(2022)Sharp, Attaiki, Crane, and Ovsjanikov]{sharp2022diffusionnet}
Nicholas Sharp, Souhaib Attaiki, Keenan Crane, and Maks Ovsjanikov.
\newblock Diffusionnet: Discretization agnostic learning on surfaces.
\newblock \emph{ACM Transactions on Graphics (TOG)}, 41\penalty0 (3):\penalty0 1--16, 2022.

\bibitem[Song and Ermon(2019)]{song2019generative}
Yang Song and Stefano Ermon.
\newblock Generative modeling by estimating gradients of the data distribution.
\newblock \emph{Advances in neural information processing systems}, 32, 2019.

\bibitem[Song et~al.(2021)Song, Sohl-Dickstein, Kingma, Kumar, Ermon, and Poole]{song2021scorebased}
Yang Song, Jascha Sohl-Dickstein, Diederik~P Kingma, Abhishek Kumar, Stefano Ermon, and Ben Poole.
\newblock Score-based generative modeling through stochastic differential equations.
\newblock In \emph{International Conference on Learning Representations}, 2021.

\bibitem[Song et~al.(2022)Song, Shen, Xing, and Ermon]{song2022solving}
Yang Song, Liyue Shen, Lei Xing, and Stefano Ermon.
\newblock Solving inverse problems in medical imaging with score-based generative models.
\newblock In \emph{International Conference on Learning Representations}, 2022.

\bibitem[Sorkine and Alexa(2007)]{sorkine2007rigid}
Olga Sorkine and Marc Alexa.
\newblock As-rigid-as-possible surface modeling.
\newblock In \emph{Symposium on Geometry processing}, pages 109--116. Citeseer, 2007.

\bibitem[Sun et~al.(2023)Sun, Mao, Jiang, Ovsjanikov, and Huang]{sun2023spatially}
Mingze Sun, Shiwei Mao, Puhua Jiang, Maks Ovsjanikov, and Ruqi Huang.
\newblock Spatially and spectrally consistent deep functional maps.
\newblock In \emph{Proceedings of the IEEE/CVF International Conference on Computer Vision}, pages 14497--14507, 2023.

\bibitem[Trappolini et~al.(2021)Trappolini, Cosmo, Moschella, Marin, Melzi, and Rodol{\`a}]{trappolini2021shape}
Giovanni Trappolini, Luca Cosmo, Luca Moschella, Riccardo Marin, Simone Melzi, and Emanuele Rodol{\`a}.
\newblock Shape registration in the time of transformers.
\newblock \emph{Advances in Neural Information Processing Systems}, 34:\penalty0 5731--5744, 2021.

\bibitem[Vincent(2011)]{vincent2011connection}
Pascal Vincent.
\newblock A connection between score matching and denoising autoencoders.
\newblock \emph{Neural computation}, 23\penalty0 (7):\penalty0 1661--1674, 2011.

\bibitem[Wang et~al.(2024)Wang, Lu, Wang, Bao, Li, Su, and Zhu]{wang2024prolificdreamer}
Zhengyi Wang, Cheng Lu, Yikai Wang, Fan Bao, Chongxuan Li, Hang Su, and Jun Zhu.
\newblock Prolificdreamer: High-fidelity and diverse text-to-3d generation with variational score distillation.
\newblock \emph{Advances in Neural Information Processing Systems}, 36, 2024.

\bibitem[Weber et~al.(2024)Weber, Dages, Gao, and Cremers]{weber2024finsler}
Simon Weber, Thomas Dages, Maolin Gao, and Daniel Cremers.
\newblock Finsler-laplace-beltrami operators with application to shape analysis.
\newblock In \emph{Proceedings of the IEEE/CVF Conference on Computer Vision and Pattern Recognition}, pages 3131--3140, 2024.

\bibitem[Wu et~al.(2024)Wu, Mildenhall, Henzler, Park, Gao, Watson, Srinivasan, Verbin, Barron, Poole, and Holynski]{Wu_2024_CVPR}
Rundi Wu, Ben Mildenhall, Philipp Henzler, Keunhong Park, Ruiqi Gao, Daniel Watson, Pratul~P. Srinivasan, Dor Verbin, Jonathan~T. Barron, Ben Poole, and Aleksander Holynski.
\newblock Reconfusion: 3d reconstruction with diffusion priors.
\newblock In \emph{Proceedings of the IEEE/CVF Conference on Computer Vision and Pattern Recognition (CVPR)}, pages 21551--21561, 2024.

\bibitem[Yan et~al.(2014)Yan, Bao, Zhang, and Wonka]{yan2014low}
Dong-Ming Yan, Guanbo Bao, Xiaopeng Zhang, and Peter Wonka.
\newblock Low-resolution remeshing using the localized restricted voronoi diagram.
\newblock \emph{IEEE Transactions on Visualization and Computer Graphics (TVCG)}, 2014.

\bibitem[Yang et~al.(2024)Yang, Huang, Sun, Bajaj, and Huang]{yang2024gencorres}
Haitao Yang, Xiangru Huang, Bo Sun, Chandrajit~L. Bajaj, and Qixing Huang.
\newblock Gencorres: Consistent shape matching via coupled implicit-explicit shape generative models.
\newblock In \emph{The Twelfth International Conference on Learning Representations}, 2024.

\bibitem[Yona et~al.(2025)Yona, Velich, Rivlin, and Kimmel]{yona2025neural}
Gal Yona, Roy Velich, Ehud Rivlin, and Ron Kimmel.
\newblock Neural descriptors: Self-supervised learning of robust local surface descriptors using polynomial patches.
\newblock In \emph{International Conference on Scale Space and Variational Methods in Computer Vision}, pages 218--230. Springer, 2025.

\bibitem[Zhu et~al.(2024)Zhu, Zhuang, and Koyejo]{zhu2024hifa}
Junzhe Zhu, Peiye Zhuang, and Sanmi Koyejo.
\newblock {HIFA}: High-fidelity text-to-3d generation with advanced diffusion guidance.
\newblock In \emph{The Twelfth International Conference on Learning Representations}, 2024.

\bibitem[Zhuravlev et~al.(2025)Zhuravlev, L{\"a}hner, and Golyanik]{zhuravlev2025denoising}
Aleksei Zhuravlev, Zorah L{\"a}hner, and Vladislav Golyanik.
\newblock Denoising functional maps: Diffusion models for shape correspondence.
\newblock In \emph{Proceedings of the Computer Vision and Pattern Recognition Conference}, pages 26899--26909, 2025.

\bibitem[Zuffi et~al.(2017)Zuffi, Kanazawa, Jacobs, and Black]{zuffi20173d}
Silvia Zuffi, Angjoo Kanazawa, David~W Jacobs, and Michael~J Black.
\newblock 3d menagerie: Modeling the 3d shape and pose of animals.
\newblock In \emph{Proceedings of the IEEE conference on computer vision and pattern recognition}, pages 6365--6373, 2017.

\end{thebibliography}
}

\maketitlesupplementary

In this supplementary material, we first provide insights on the sign ambiguity problem of functional maps in~\cref{sec:signamb}. We provide more experimentation details about datasets (\cref{sec:dsets}), baselines (\cref{sec:base}), and implementation (\cref{sec:exps}) for the experiments in Sec.~5 of the paper. Next, we show the behavior of our method with a plot of the loss during zero-shot optimization in \cref{sec:loss}, a visualization of denoising trajectories in \cref{sec:trajiiis}, and finally an analysis of descriptors in \cref{sec:dedescs}.  We also show that the matching provided by our method allows competitive reconstruction of input shapes by combining it with the ARAP energy in \cref{sec:rsescs}. Finally, in \cref{sec:therock}, we provide a simple experiment providing insights on sparsity-promoting mask efficiency.

\section{Sign Ambiguity of Functional Maps}\label{sec:signamb}
\begin{figure}
    \centering
    \includegraphics[width=0.95\linewidth]{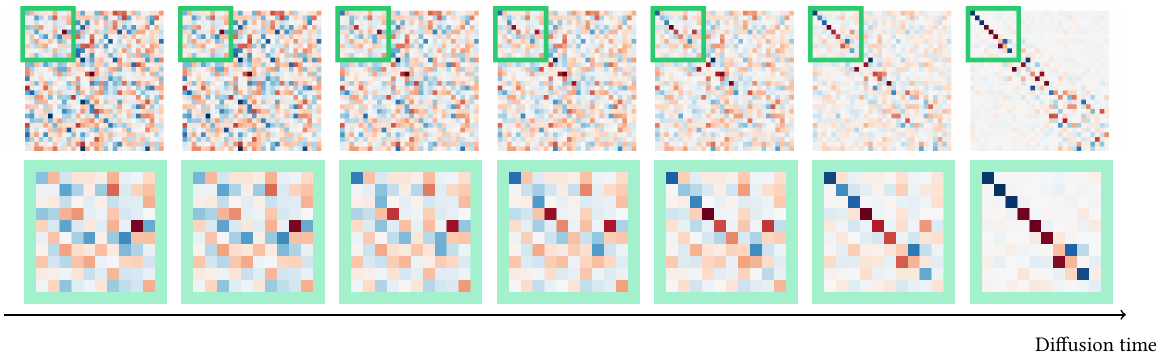}
    \caption{Generation process of a functional map using a diffusion model. For low-frequency elements (green square), the sign of diagonal elements at the Gaussian noise step never changes during the denoising process. This explains why spectral regularization with SDS fails to correct misalignments effectively.}
    \label{fig:sign}
\end{figure}

This phenomenon occurs because functional maps are nearly discrete at low frequencies. Indeed, it has been observed that the ground truth maps at low frequency follow a diagonal structure~\cite{funcmaps}, where the values of the diagonal elements are $\pm 1$ (modulo volume changes). This affects the overall trajectory of generation - where signs of the diagonal elements remain unchanged (\cref{fig:sign}) - and thus the capacity of diffusion models to provide efficient spectral regularization. Thus, to better capture the underlying structure of the functional maps from data, we chose to adopt a \textbf{sign-agnostic} approach.

\section{Datasets}~\label{sec:dsets}

\myparagraph{Near-isometric Shape Matching.} The FAUST remeshed~\cite{ren2018continuous, bogo2014faust} version contains 10 individuals in 10 different poses. SCAPE~\cite{anguelov2005scape} contains 50 challenging poses of one individual. SHREC~\cite{melzi2019shrec} contains 50 humans from different datasets, with 407 annotated pairs using an automatic human registration algorithm (partial shape matching pairs are excluded).

\myparagraph{Non-isometric Shape Matching.} The matching version of the DT4D dataset~\cite{magnet2022smooth} contains more than 400 shapes, with more than 1000 annotated pairs remeshed using the LRVD algorithm~\cite{yan2014low}, from which we use the intra-category and inter-category test sets from~\cite{li2022learning}. The SMAL remeshed dataset~\cite{donati2022complex}, which contains around 400 animal pairs extracted from real images using the SMAL deformation model~\cite{zuffi20173d}. The animal shape pairs from the TOSCA are from \textit{cat, dog, horse} and \textit{wolf} categories.

\section{Baselines}~\label{sec:base}

We compare our method against several baselines for shape matching. 3D-CODED~\cite{groueix2018_3DCODED} is an autoencoder trained specifically for shape matching. The shape latent vectors are computed and refined by optimizing the obtained registrations. Neural Jacobian Fields~\cite{NJF_2022} is a model that predicts the Jacobian of deformation instead of vertex positions and generalizes to unregistered meshes. Smooth shells~\cite{eisenberger2020smooth} is an axiomatic approach that refines functional maps in a coarse-to-fine approach to obtain plausible final correspondences. Shape-Non-Rigid-Kinematics (SNK)~\cite{attaiki2023shape} is a state-of-the-art zero-shot algorithm to train deep feature extractors on pairs of shapes. We also compare to a state-of-the-art deep functional maps approach, Simplified Fmaps~\cite{magnet2024memory}. All trainable models are trained on the D-FAUST dataset. Finally, we also show the results of using a feature extractor with random weights combined with different masks.

\section{Experimental Details}\label{sec:exps}
\myparagraph{Feature Extractor.} We follow the zero-shot experimental settings from SNK~\cite{attaiki2023shape}. The feature extractor consists of four DiffusionNet blocks of dimension 256, and we use 128 eigenvectors for the heat diffusion. The input features of the feature extractor are XYZ features on the oriented versions of each dataset~\cite{sharma2020weakly}. We set $\lambda=0.1$ for humans and $\lambda=1e-3$ for the other datasets, respectively. For the Ini+Zoomout scenario with our mask, we set $\lambda=1$.

\myparagraph{Diffusion Model Training.} We train our spectral diffusion model for 1000 steps. The training setting is the same as in~\cite{karras2022elucidating}, with optimal reweighting of the losses and using the variance-preserving SDE, which reproduces the trajectory of DDPM~\cite{ho2020denoising}. No normalization of functional maps is applied, as the values inside the matrices range from -1 to 1 already.

\myparagraph{Zero-Shot Training.} We train our deep functional map approach for 1000 gradient steps using Adam optimizer. The overall training on a single pair takes approximately 180 seconds on a NVIDIA L40S GPU. 

\myparagraph{Evaluation.} For the evaluation, we refined our optimized maps using Zoomout to obtain a final map dimension of 150x150, as commonly done in the deep functional maps approach~\cite{attaiki2023shape, magnet2024memory}.   

\section{Loss Behavior}\label{sec:loss}
\begin{figure}
    \centering
    \includegraphics[width=0.9\linewidth]{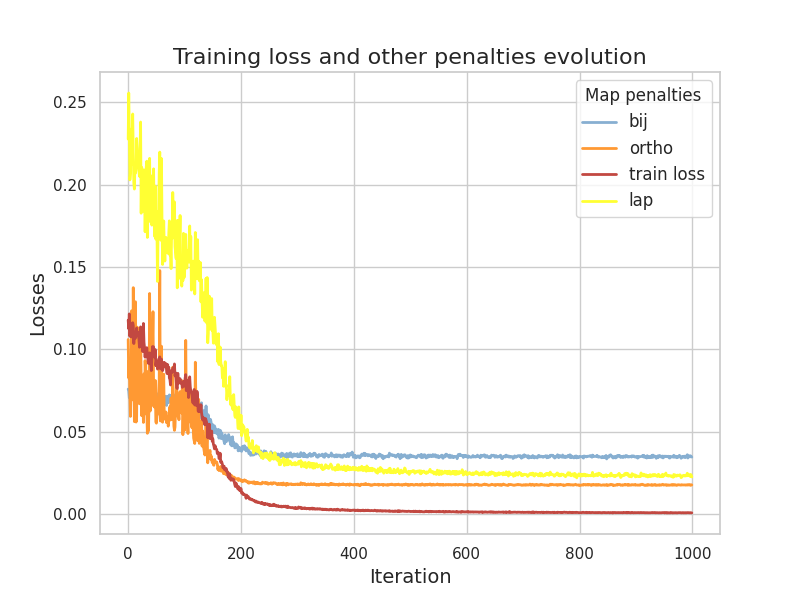}
    \caption{Loss and other penalties during optimization of the matching.}
    \label{fig:enter-loss}
\end{figure}

We plot the loss behavior during optimization in~\Cref{fig:enter-loss}. The loss is smoothly optimized and converges rapidly.

\section{Generating Functional Maps and Absolute Functional Maps}\label{sec:trajiiis}

\begin{figure}
    \centering
    \includegraphics[width=0.95\linewidth]{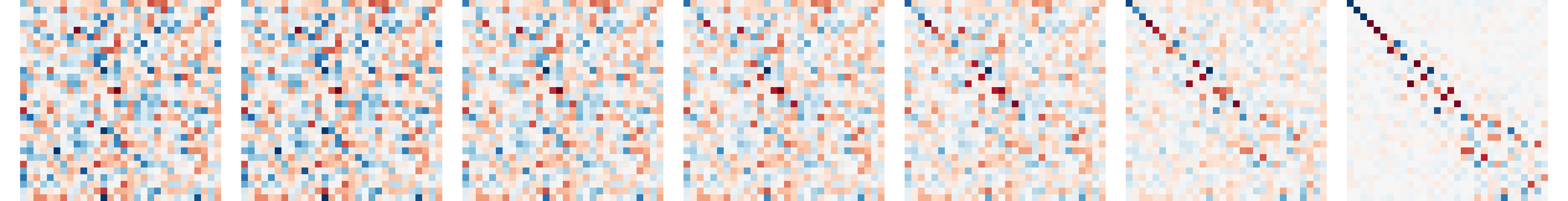}
    
    \includegraphics[width=0.95\linewidth]{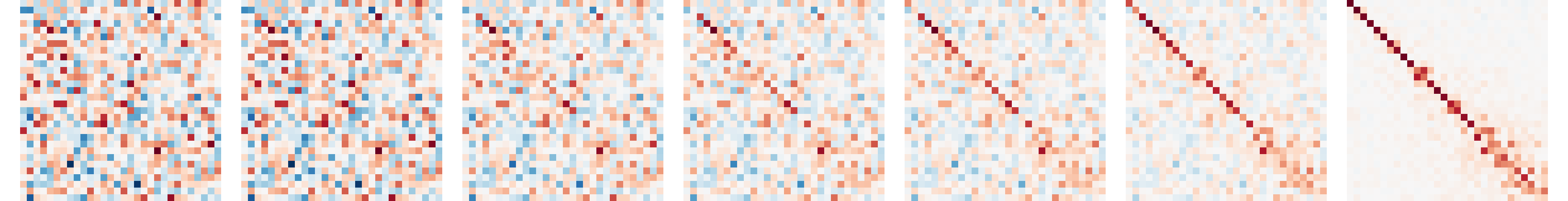}
    \caption{Example generation trajectories using spectral diffusion models on functional maps (top) and absolute functional maps (bottom)}
    \label{fig:trajs}
\end{figure}

We show two example denoising trajectories, from the original and absolute spectral diffusion models in~\Cref{fig:trajs}.

\section{Quality of Learned Descriptors}\label{sec:dedescs}

\begin{figure}
    \centering
    \includegraphics[width=0.95\linewidth]{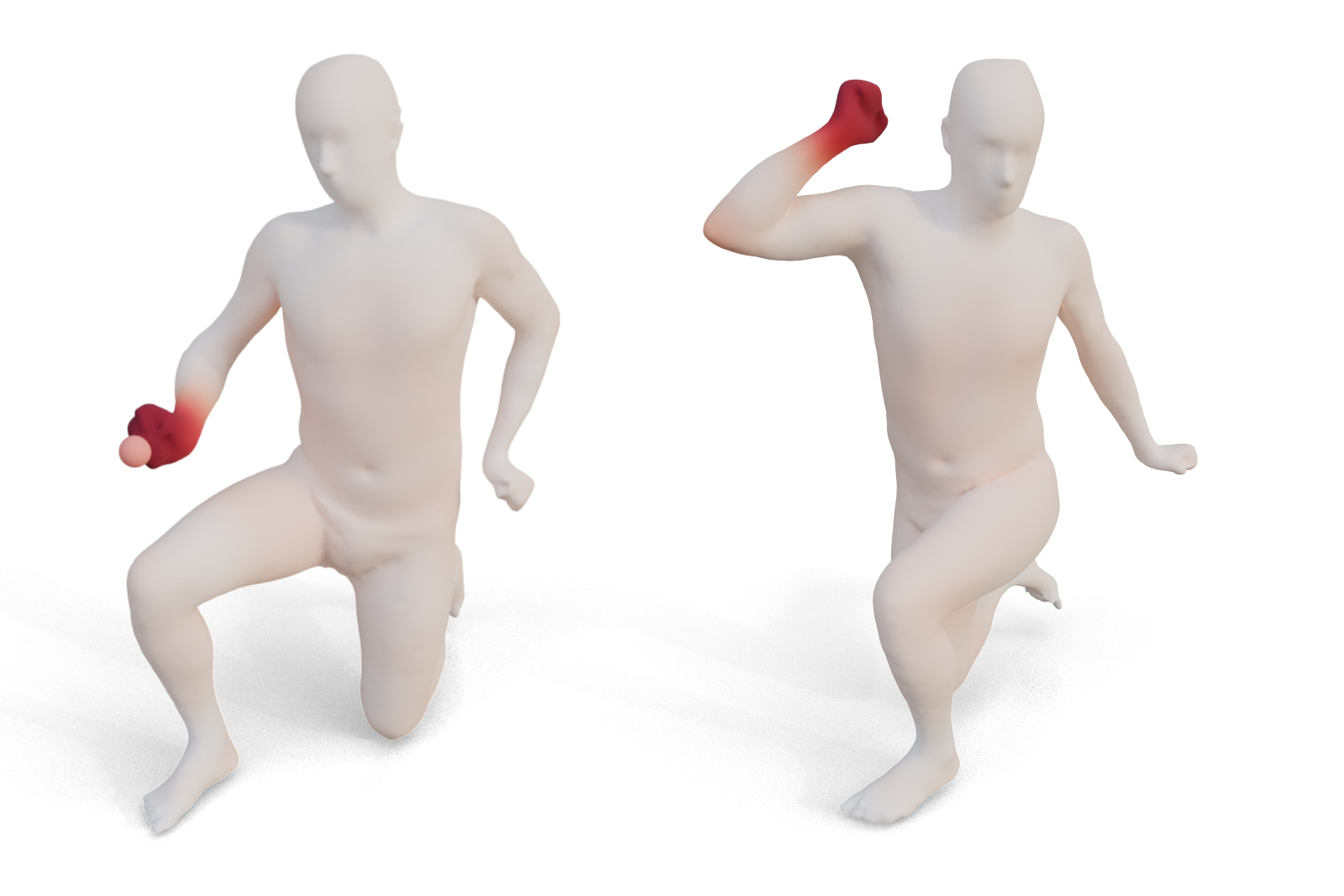}
    \caption{After applying DiffuMatch, we select a point on the source shape and compute the distance of this point to all points on both target and source shapes, in the descriptor space. We plot the obtained distances on both shapes. The closest points are points that are geodesically close to the select point.}
    \label{fig:pointwise}
\end{figure}

Learned descriptors using our approach are meaningful thanks to our proper loss. Indeed, it has been shown that when properness is encouraged, the extracted correspondence is approximately the same whether it is extracted from the functional map or by nearest neighbor search~\cite{attaiki2023understanding}. We visually verify this in~\Cref{fig:pointwise}, where we show the nearest points to a selected point using nearest neighbor in the feature space (after projection on the space spanned by the first 30 eigenfunctions -- the only ones used in the map computation), showing that our method enables meaningful descriptor learning in addition to the quality of the shape matching.

\section{Comparison of Reconstruction of Deformation Models}\label{sec:rsescs}

As stated in the paper, deformation models are not suitable for generalization to new type of categories. In this section, we provide reconstructions from 3D-CODED and NJF of the source shape in section 4.2. Moreover, as SNK provides a shape reconstruction as output, we also show the reconstruction provided by SNK. Finally, we extract shape correspondence $\Pi$ from DiffuMatch and reconstruct the vertex position of the shape in the target mesh topology, by solving for the closest possible solution minimizing the As-Rigid-As-Possible (ARAP)~\cite{sorkine2007rigid}. Let $X$ be the vertex of the source mesh, the reconstruction $Y_{rec}$ in the target mesh topology is given by:
$$
Y_{rec} = \underset{Y}{\text{argmin}} ||Y - \Pi X||² + E_{arap}(Y).
$$

\begin{figure}
    \centering
    \includegraphics[width=0.95\linewidth]{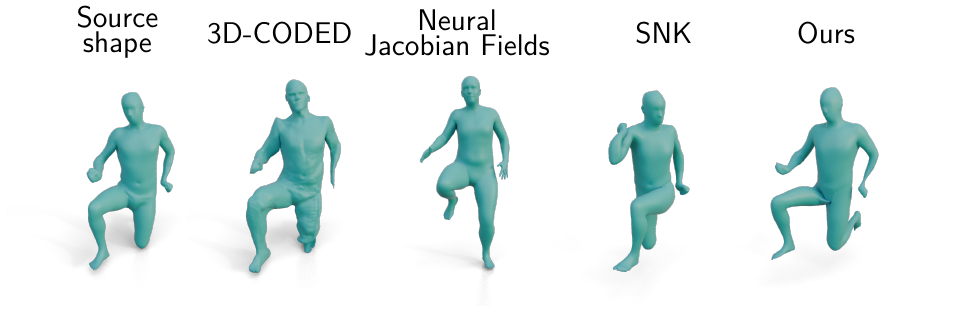}
    \caption{Reconstruction of source shape using different approaches. For our reconstruction, we solve for the closest vertex positions to the matched shape minimizing the ARAP energy from the target shape.}
    \label{fig:arap}
\end{figure}

As our matching is nearly perfect, the provided reconstruction, shown in~\Cref{fig:arap} is visually better than the one given by other approaches, up to some artifacts due to our matching being computing on the first 30 eigenfunctions only. The capabilities of our model can also be extended to reconstruction of input meshes in a new topologies.

\section{Importance of Mask Regularization.}\label{sec:therock}
\begin{figure}
    \centering
    \includegraphics[width=0.8\linewidth]{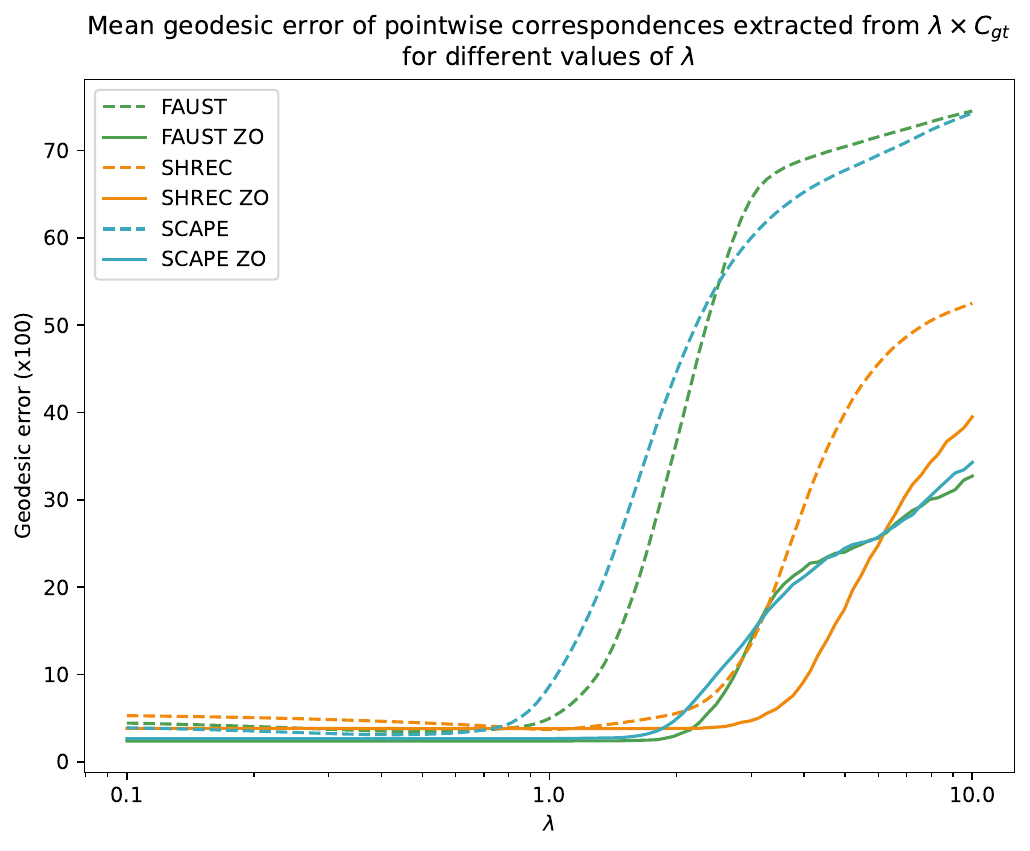}
    \caption{Given a ground truth functional map $C \in \mathcal{M}_n(\mathbb{R})$, and a scalar, $0< \lambda < 1$, the matrix $\lambda C$ represents approximately the same pointwise correspondence as C. By applying Zoomout to both $C$ and $\lambda C$, we obtain the same map. The observation does not always hold when $lambda>1$. We plot the geodesic errors of $\lambda C$ for different values of $\lambda$.}
    \label{fig:exp_lambda}
\end{figure}

Mask regularization plays a key role in most (deep) functional map pipelines~\cite{attaiki2022ncp}. We run a simple experiment to show that the functional map space is particularly well-suited for this type of penalty. Multiplying a ground truth functional map matrix $C \in \mathcal{M}_n(\mathbb{R})$ by any scalar $0<\lambda < 1$ raises approximately the same pointwise correspondence as the original one from $C$. We also observed the same phenomena after applying Zoomout~\cite{zoomout}, where the obtained correspondences are the same. This phenomenon is illustrated in~\Cref{fig:exp_lambda}.

As most masks are sparsity-promoting masks, their mask penalty minimizers have multiple solutions, which are $\lambda \times X$ where $X$ is any solution. As we observed, optimized maps can be proportional to the ground truth solution and still output a correct pointwise correspondence. Based on this insight and the efficiency of mask regularization in functional map computation, we proposed to distill the knowledge of our trained diffusion model by extracting a sign-agnostic mask that will promote structures seen in the training set.

\section{Computation Time}
A single run of DiffuMatch takes approximately 150 seconds on an NVIDIA L40S GPU. 
In the case where computation time is a bottleneck, the scenario Ini (feature extractor with random weights) + Zoomout \textit{with our distilled mask} is competitive as it requires little computation time. We provide a comparison of computation time with some other competing methods in~\cref{tab:cost_train}
\begin{table}[]
    \centering
    \begin{tabular}{l|c}
    Method & Computation time \\ \hline \hline 
    3D-Coded & 160s \\ \hline
    Neural Jacobian Fields & 3.26s \\ \hline 
    SimplifiedFmaps & 1.08s \\ \hline 
    SNK & 130s \\ \hline \hline 
    Ini + Zoomout (our mask) & 0.75s \\ \hline 
    Ours full & 150s \\ 
    \end{tabular}
    \caption{Computation costs for different methods.}
    \label{tab:cost_train}
\end{table}

\begin{figure}
    \centering
    \scriptsize
    \includegraphics[width=0.7\linewidth]{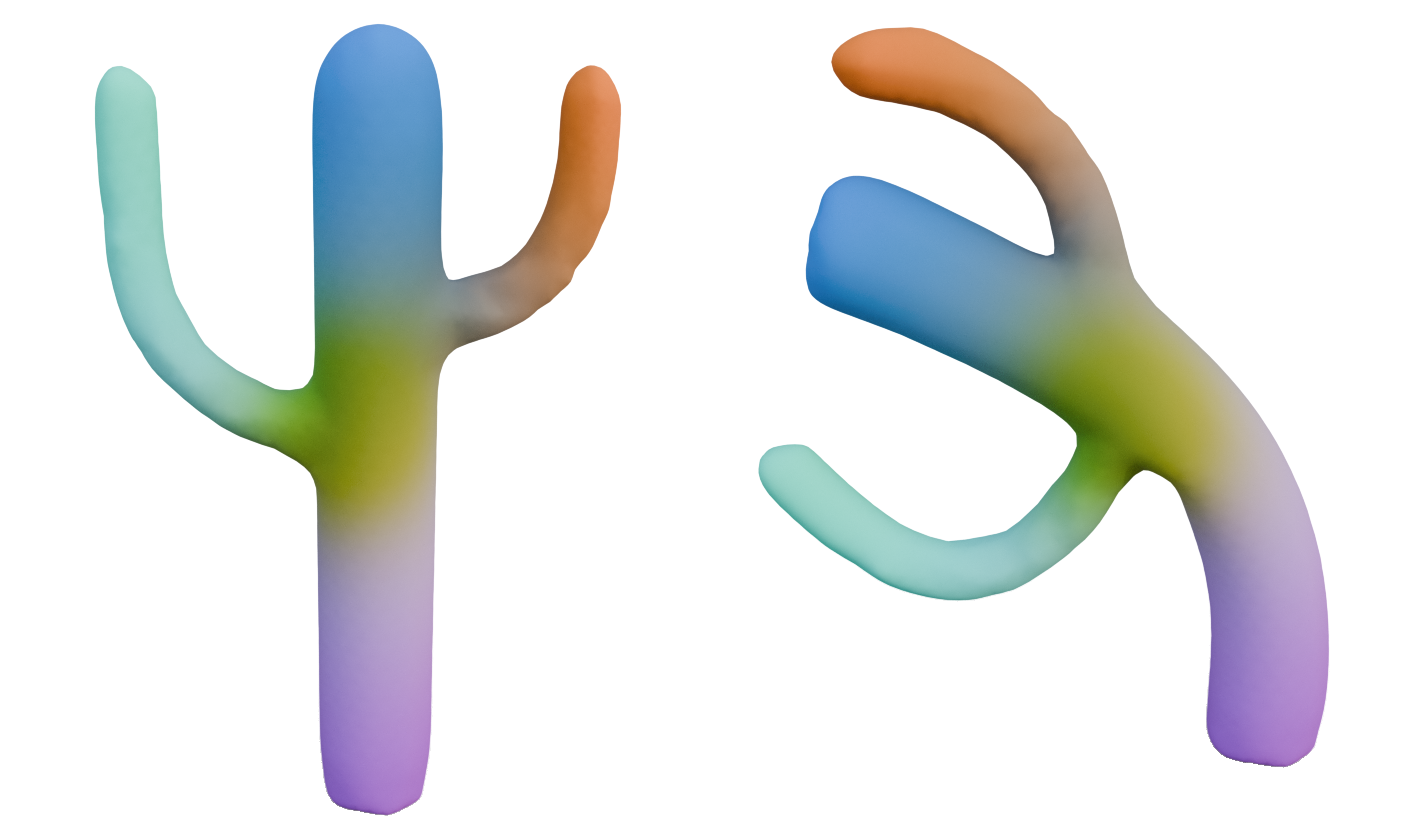}
    \caption{DiffuMatch result on a cactus pair.}
    \label{fig:cactus}
\end{figure}

\begin{figure}
    \centering
    \includegraphics[width=0.8\linewidth]{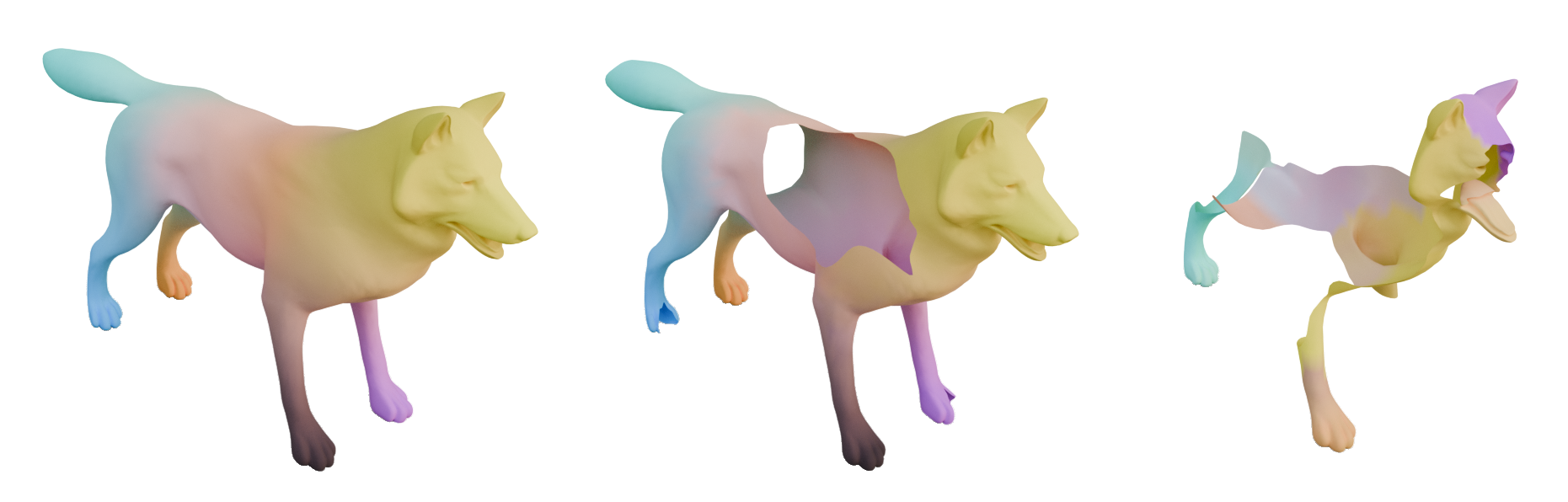}
    \caption{Partial matching results on SHREC16}
    \label{fig:partial}
\end{figure}

\section{Generalization}

\noindent {\bf Non articulated shapes}  We showcase that DiffuMatch can perform well on a pair of two cactus meshes in ~\cref{fig:cactus}.

\noindent {\bf Partial shape matching} We show in Fig~\ref{fig:partial} some partial matching results. DiffuMatch can work on pairs where the partiality is moderate. However, when the partiality becomes significant, DiffuMatch is prone to failure, with an error of $19.8$ and $23.4$ on SHREC16 cuts and holes partial shape matching challenges~\cite{rodola2017partial}. This is to be expected, as functional maps have a different structure between full and partial correspondence~\cite{rodola2017partial}, and methods applied to partial shape matching often rely on modified losses ~\cite{cao2022unsupervised, Cao2023UnsupervisedLO} or require feature pre-training~\cite{Cao2023UnsupervisedLO}.

\end{document}